\documentclass{article}
\usepackage[table]{xcolor}
\usepackage{hyperref}
\usepackage{enumitem}
\usepackage{arxiv}
\usepackage[utf8]{inputenc} 
\usepackage[T1]{fontenc}    
\usepackage{hyperref}       
\usepackage{url}            
\usepackage{booktabs}       
\usepackage{amsfonts}       
\usepackage{nicefrac}       
\usepackage{microtype}      
\usepackage{lipsum}
\usepackage{graphicx}
\usepackage{amssymb}
\usepackage{latexsym}
\usepackage{url}
\usepackage{xcolor}
\definecolor{newcolor}{rgb}{.8,.349,.1}
\usepackage{algorithm}
\usepackage{algorithmicx}
\usepackage[noend]{algpseudocode}
\usepackage{fancyhdr}
\usepackage{graphicx}
\usepackage[noend]{algpseudocode}
\usepackage{fancyhdr}
\usepackage{subfigure}
\usepackage[cmex10]{amsmath}
\usepackage{amsfonts}
\usepackage{wrapfig}
\usepackage{color}
\usepackage{verbatim}  
\usepackage{enumitem} 
\usepackage{changepage} 
\usepackage{mdwlist}
\usepackage{paralist}
\usepackage{pdfpages}
\usepackage{graphicx}
\usepackage{float} 
\let\tabularnewline\\

\newcounter{defn}

\title{Near real-time map building with multi-class image set labelling and classification of road conditions using convolutional neural networks}

\author{
	Sheela~Ramanna\thanks{Corresponding author. This research has been supported by the Natural Sciences and Engineering Research Council of Canada (NSERC) Engage grant EGP 537911-18.} \\
	Department of Applied Computer Science\\ University of Winnipeg,\\
	Winnipeg, Manitoba R3B 2E9, Canada\\
	\texttt{s.ramanna@uwinnipeg.ca} \\
	\And
	Cenker~Sengoz\\
	Department of Applied Computer Science\\ University of Winnipeg,\\
	Winnipeg, Manitoba R3B 2E9, Canada\\
	\texttt{cenker.sengoz@gmail.com} \\
	\And
	Scott~Kehler\\
	Weatherlogics Inc.\\ 441-100 Innovation Drive,\\
	Winnipeg, Manitoba R3T 6G2, Canada\\
	\texttt{scott@weatherlogics.com} \\
	\And
	Dat~Pham\\
	Department of Applied Computer Science\\ University of Winnipeg,\\
	Winnipeg, Manitoba R3B 2E9, Canada\\
	\texttt{phdatdt@gmail.com} \\
}

\begin{document}
\maketitle

\begin{abstract}
Weather is an important factor affecting transportation and road safety.  In this paper, we leverage state-of-the-art convolutional neural networks in labelling images taken by street and highway cameras located across across North America.  Road camera snapshots were used in experiments with multiple deep learning frameworks to classify images by road condition. The training data for these experiments used images labelled as dry, wet, snow/ice, poor, and offline. The experiments tested different configurations of six convolutional neural networks (VGG-16, ResNet50, Xception, InceptionResNetV2, EfficientNet-B0 and EfficientNet-B4)  to assess their suitability to this problem. The precision, accuracy, and recall were measured for each framework configuration. In addition, the training sets were varied both in overall size and by size of individual classes. The final training set included 47,000 images labelled using the five aforementioned classes. The EfficientNet-B4 framework was found to be most suitable to this problem, achieving validation accuracy of 90.6\%, although EfficientNet-B0 achieved an accuracy of 90.3\% with half the execution time. It was observed that VGG-16 with transfer learning proved to be very useful for data acquisition and pseudo-labelling with limited hardware resources, throughout this project. The EfficientNet-B4 framework was then placed into a real-time production environment, where images could be classified in real-time on an ongoing basis. The classified images were then used to construct a map showing real-time road conditions at various camera locations across North America.  The choice of these frameworks and our analysis take into account unique requirements of real-time map building functions. A detailed analysis of the process of semi-automated dataset labelling using these frameworks is also presented in this paper.  
\end{abstract}

\section{Introduction}
\label{sec:introduction}

Adverse road conditions present a frequent hazard to motorists. In cold climates, snow, ice, and frost can produce slippery roads, while the reduced friction from wet roads is a hazard in both warm and cold climates. Data from 2010-2018 in the United States showed that on average 767,779 crashes per year (13\% of the total) occur during adverse weather conditions (rain, snow, sleet, freezing rain, or hail). In addition, an average of 2,747 fatalities per year (9\% of all fatalities) occurred during times of adverse weather conditions\footnote{https://cdan.dot.gov/query, retrieved on Jan. 4, 2020}. Advances have been made in better monitoring roads during hazardous weather conditions. Road Weather Information Systems (RWIS) can provide real-time road weather information at point locations, which is often used to produce road weather forecast(e.g.~\cite{crevier2001,Sass1997}). This data is then transmitted to transportation operations centers and disseminated to the public through services such as the 511 network~\cite{Drobot2014}. Many RWIS are also equipped with cameras, which give a real-time view of the road. While the information provided by RWIS and cameras is useful, there are still limitations. Since these systems are operated at the state/province or local level, there is no unified road information system. Therefore, motorists must consult different sources for road weather information in each jurisdictions where they travel. Due to the cost of such systems, not all jurisdictions have RWIS/cameras and those that do often have a limited number. This can introduce large gaps in road weather information. These gaps are sometimes filled by manual observations from operators, or have no data at all.
Since cameras are much more prevalent than RWIS, and less expensive, they may present an opportunity to improve road weather data where there is currently limited data. In addition, since camera images are readily available and come in common formats, they can be sourced across all jurisdictions and combined into a unified system. However, combining all such cameras is a major task, since there are tens of thousands in North America alone. Furthermore, as noted by Carrillo et al.~\cite{Carillo2019} it is challenging for operators to process vast amounts of road weather data in real time.

Early work involving road condition classification from weather data involved cameras mounted on vehicles and primarily used in vehicle navigation~\cite{KurihataRain2005,Hautiere2006,RoserRain2008,Yan2009,Bronte2009FogDS,Omer2010AnAI, Gallen2011,Pavli2012,Zhang2015MulticlassWC,Almazan2016}.   Some of these methods use  image processing techniques such as extracting regions of interest (ROI) from the images or road segmentation.  Histogram features derived from the ROIs can then be used with classical machine learning methods such as Support Vector Machines (SVM) to label the weather/road conditions into various categories such as sunny, cloudy and rainy.   Road condition estimation based on this spatio-temporal approach to model wet road surface conditions, that integrates over many frames, was explored~\cite{Amtor2015}. Weather recognition from general outdoor images was explored in~\cite{Narasimhan2003,LICVPR2009,Laffont2014,Hong2014,Li2014} to name a few.  

The success of deep convolutional neural networks(DCNN)~\cite{schmidhuber2015deep,lecun2015deep}, in computer vision tasks~\cite{Krizhevsky2012,Russakovsky2015} and the generation of large weather datasets~\cite{LiuPAMI2017,RSCMLin2017,zhao2019cnnrnn}, led to their application in weather recognition problems~\cite{Elhoseiny2015WeatherCW,Zhu2016ExtremeWR,RSCMLin2017,Guerra2018,LiGAN2018,zhao2019cnnrnn}.   Automatic fog detection with DCNNs using  H20 platform\footnote{https://www.h2o.ai/} 
to predict the presence of dense fog from daytime camera images has been implemented with several sets of images collected by Royal Netherlands Meteorological Institute(KNMI)\footnote{https://www.wmo.int/pages/prog/www/IMOP/documents/O4\_1\_Pagani\_etal\_ExtendedAbstract.pdf}.  A near-real-time geographical map showing the predicted values of the cameras is also given with promising results. In~\cite{Pan2018RoadSurface}, 5000 images from highway sections in Ontario, Canada captured by smartphones were used to classify road surface conditions using a pre-trained VGG-16 model.  For a five-class road surface classification, the best accuracy with DCNN was 78.5\%. In~\cite{Nolte2018RoadSurface}, two DCNNS were applied to differentiate six classes of road surface conditions such as cobblestone, wet asphalt, snow, grass, dirt and asphalt with an eventual goal of predicting the road friction coefficient.  This study augmented their dataset with data from publicly available datasets for automated driving, which led to a classification accuracy of 92\%.  Classification of road surface condition with deep learning models was explored by Carillo et al.~\cite{Carillo2019} where six state-of-the-art DCNN models were pre-trained using ImageNet parameters to classify to road surface condition images (about 16,800) from roadside cameras in Ontario, Canada. 

\begin{figure}[!ht]
	\centering
	\centerline{
		\subfigure[]{\includegraphics[width=25mm]{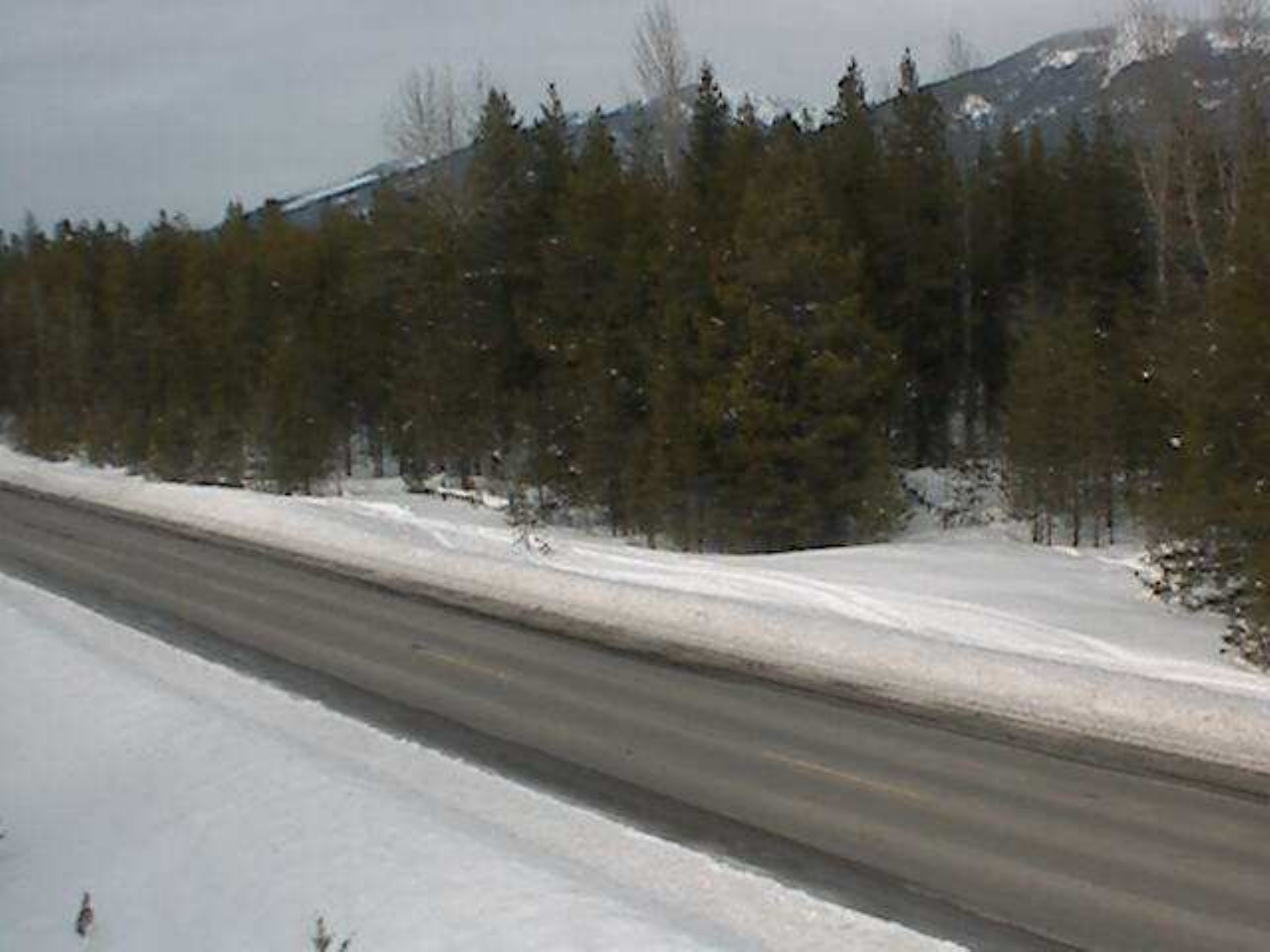}\label{fig:dry12}}
		\hspace{1mm}
		\subfigure[]{\includegraphics[width=25mm]{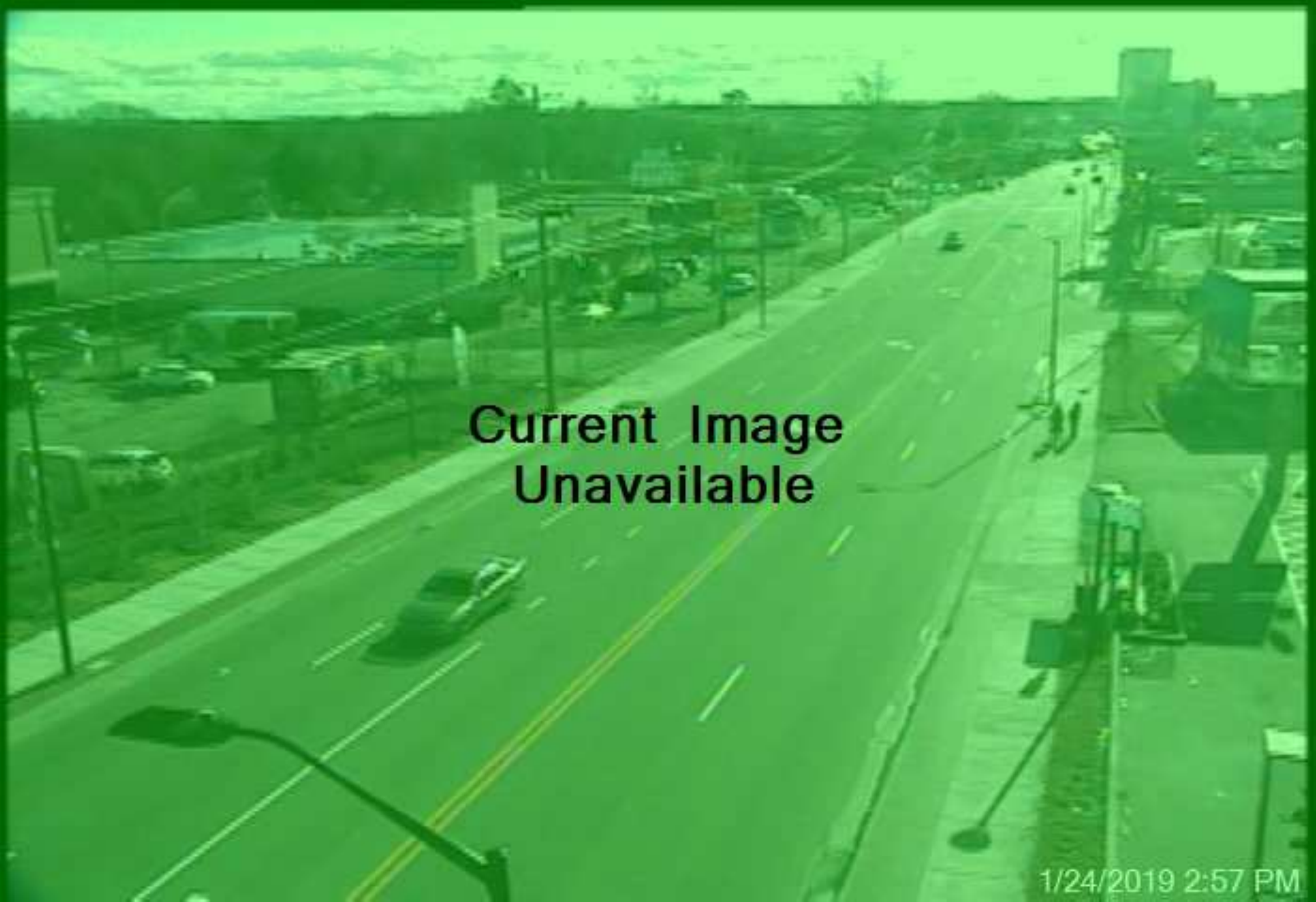}\label{fig:off10}}
		\hspace{1mm}
		\subfigure[]{\includegraphics[width=25mm]{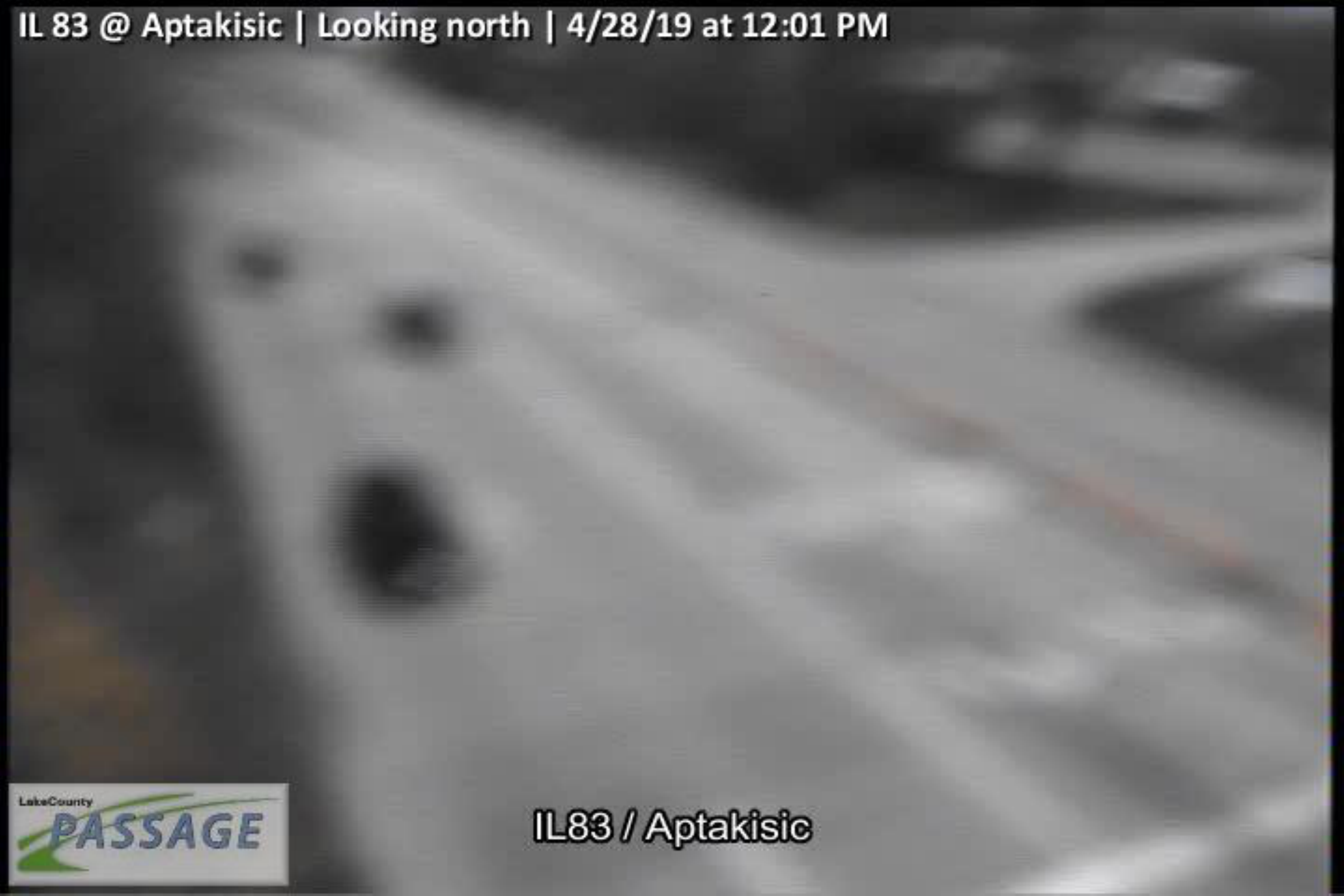}\label{fig:poor8}}
			\hspace{1mm}
		\subfigure[]{\includegraphics[width=25mm]{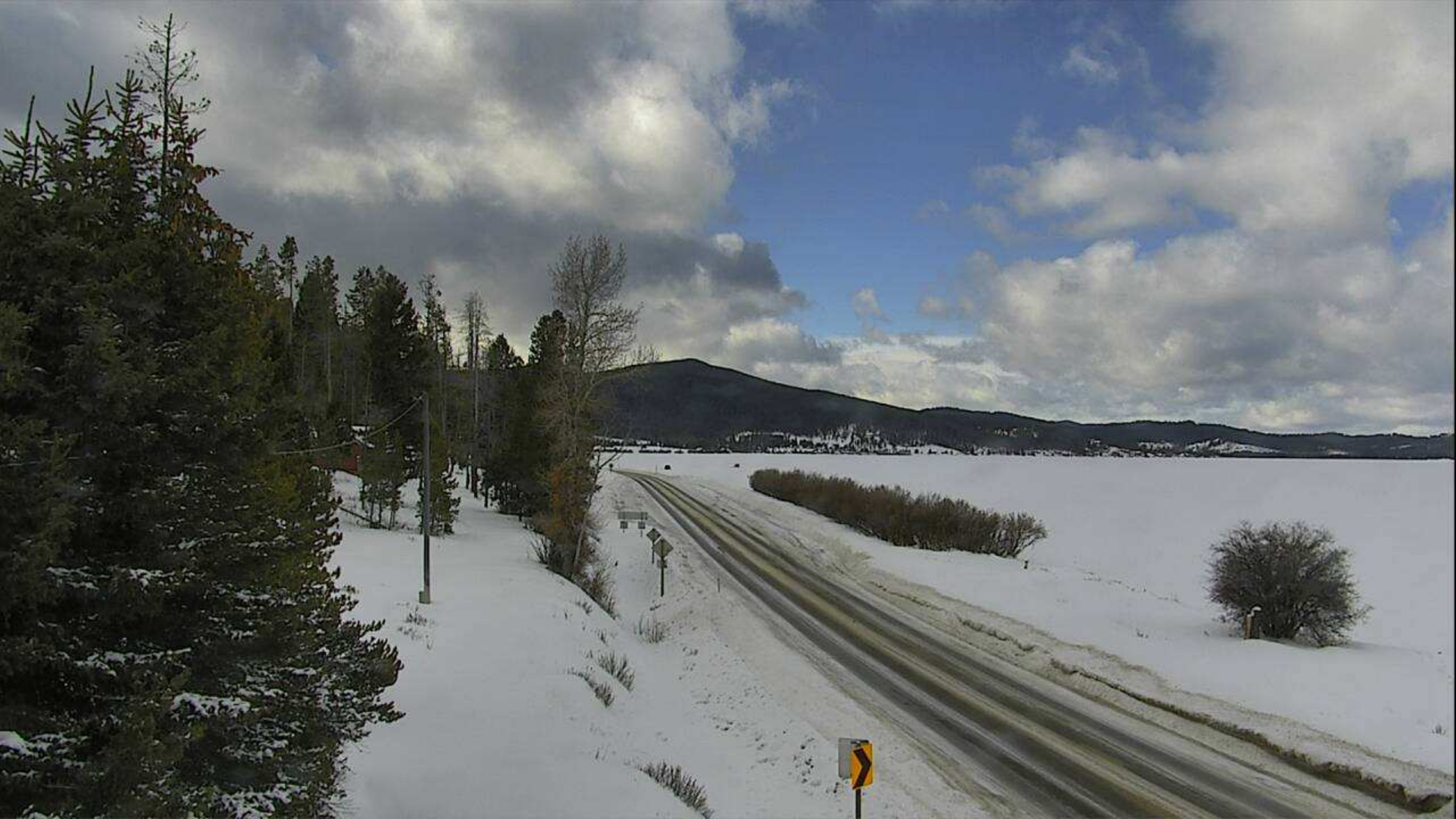}\label{fig:snow8}}
				\hspace{1mm}
		\subfigure[]{\includegraphics[width=25mm]{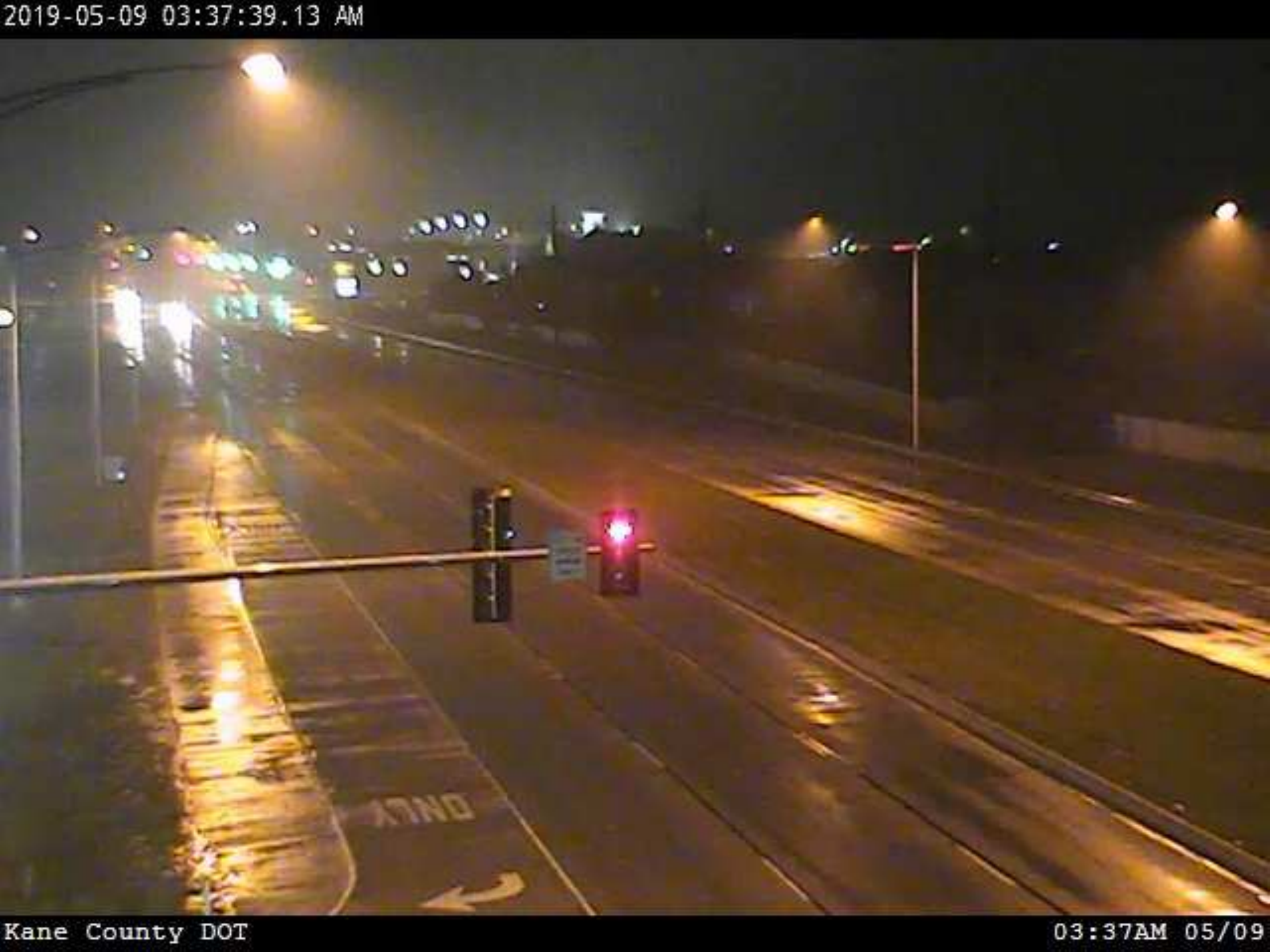}\label{fig:wet11}}
	}
	\caption{Sample raw images from street and road cameras representing (a) Dry (b) Offline, (c) Poor, (d) snow and (g) wet categories}
	\label{fig:camera}
\end{figure}

The research goals explored in this paper are as follows:  i) to leverage state-of the art DCNN's in labelling images taken by street and highway cameras located across Canada and the United States (see Figure~\ref{fig:camera}),  ii) to evaluate multiple DCNN models for classification of road conditions, and iii) to construct a real-time map of North-America depicting road conditions.

%

The  major contributions of this work are as follows: i) a methodology outlining steps for generating a multi-class dataset of images of road conditions in North America from road cameras, ii) a map building system, and  iii) detailed analysis of the process of semi-automated dataset labelling using well-known deep learning frameworks. It is also noteworthy that the best classification accuracy  result of 90.9\% was achieved without performing any pre-processing on the images (such as noise removal, text/logo removal, histogram equalization, cropping) other than rescaling. 

Our paper is organized as follows: In Section~\ref{sec:related_works}, we briefly review  research related to weather classification.  In Section~\ref{sec:pipeline}, an overview of the map building application pipeline.  In addition, we give an overview
of the various convolutional network architectures used in this paper.  In Section~\ref{sec:method}, we present a detailed discussion of the process of labelling raw images and generating training examples.  In Section~\ref{sec:finalresults}, we give classification results and illustrate the final map building exercise.  Lastly, we give concluding remarks in Section~\ref{sec:con}.

\section{Related Works}
\label{sec:related_works}
Research related to weather classification using deep learning is presented here. Related works are grouped into different categories  that share similar methods.
\vspace{3mm}
\begin{compactenum}[{\bf }1$^o$]
 \item {\bf Vehicle navigation from in-vehicle cameras using hand-crafted features}: \\
In~\cite{KurihataRain2005}, raindrops formed on the windshield of vehicles were used to detect \emph{rainy weather} where eigendrops representing the principal components were extracted from the raindrop images. In~\cite{RoserRain2008,Yan2009}, various histogram features were extracted from the captured images.  These features were used as inputs to well-known machine learning classifiers such as SVM or AdaBoost to classify the images into classes such as sunny, cloudy and rainy~\cite{Yan2009}, or more granular categories such as clear weather, light rain or heavy rain~\cite{RoserRain2008}. Another important weather condition besides rain, is fog. In~\cite{Hautiere2006}, a single camera image was captured consisting of road and sky elements, with the intention of detecting day-time fog and the computation of the meteorological visibility distance for vehicle navigation. In a later paper,~\cite{Gallen2011}, a night-fog detection system was developed using two methods that detect fog in presence of road traffic lights, or public lighting, using multipurpose cameras. In ~\cite{Bronte2009FogDS}, a real-time fog detection system was developed using a combination of image processing techniques such as Sobel filtering (to detect blurry images), road/sky segmentation and visibility distance calculation. Another day-time fog detection method  using global features in terms of the power spectrum by training different scaled and oriented Gabor filters, was presented in~\cite{Pavli2012}. A SVM classifier with an RBF kernel was used for classification. In~\cite{Zhang2015MulticlassWC}, multiple features such as sky, shadow, rain streak, snowflakes, dark channel, contrast and
saturation were extracted from  an outdoor data set consisting of 20K images.  These images were classified based on shared dictionaries of weather conditions  and a SVM classifier.
 
\item {\bf Weather recognition from outdoor images}:  \\
In~\cite{LICVPR2009}, a photometric stereo-based method was proposed to estimate weather conditions using multiple popular tourist site images from the internet. In~\cite{Narasimhan2003}, a physics-based model was developed to capture multiple scattering of light rays as they travel from a source to an observer, for various weather conditions including fog, haze, mist and rain. In~\cite{Laffont2014}, the weather recognition problem was viewed as a dynamic scene changing scenario where several transient attributes such as lighting, weather, and seasons were defined and used to annotate thousands of images with perceived scene properties. Support vector regression and logistic regression methods were used to train different non-linear predictors. In~\cite{Hong2014,Li2014}, classifiers such as k-nearest neighbour and decision trees were used to detect the weather conditions in outdoor images using global -features such as power spectral slope, edge gradient energy, contrast, saturation, and noisy images. In~\cite{Almazan2016}, the authors first segmented road surface images to obtain a ROI showing different weather conditions from bare dry to snow packed. This segmentation method relied on contextual information to define the vanishing point of the road and horizon line.  These images were then classified into five classes (dry, wet, snow, ice, packed) using a standard SVM classifier with an RBF kernel.  This method resulted in an accuracy of 86\% for binary classification (bare vs. snow or ice-covered)

 \item {\bf Weather recognition with features derived from CNNs}: \\
In~\cite{Elhoseiny2015WeatherCW}, a CNN was trained using ImageNet to classify weather images. In~\cite{LiuPAMI2017}, a collaborative learning approach using novel weather features to label a single outdoor image as either sunny or cloudy, was proposed. A CNN was used to extract features which were then fed to an SVM framework to obtain individual weather features. In addition, a data augmentation scheme was used to enrich the training data. In~\cite{RSCMLin2017}, multi-class benchmark dataset containing six common categories for sunny, cloudy, rainy, snowy, haze, and thunder weather was created. A region selection and concurrency (RSCM) was proposed to detect visual concurrency on region pairs of weather categories.  This model was tested using a deep-learning framework. In~\cite{Zhu2016ExtremeWR}, features of extreme weather and recognition models were generated from a large-scale extreme weather dataset in which 16635 extreme weather images with complex scenes were divided into four classes (sunny, rainstorm, blizzard, and fog). A pre-trained ILSVRC-2012 dataset was used in conjunction with GoogLeNet to fine-tuned their dataset.
In~\cite{Guerra2018}, an open source rain-fog-snow (RFS) dataset of images was created.  A novel algorithm where super-pixel delimiting masks as a form of data augmentation was proposed. A CNN was used to extract features from the augmented images, which were then used to train an SVM classifier. In~\cite{LiGAN2018}, deep convolution generative adversarial networks (DCGAN) were used to generate images to balance, three benchmark (imbalanced) datasets of weather images.  The CNN  model was then applied directly to classify the images. In~\cite{zhao2019cnnrnn}, weather recognition was treated as a multi-label classification task where an image was assigned to more than one label according to weather conditions. The authors also used a CNN to extract the most correlated visual weather features. A  long-short-term-memory version of the recurrent neural network (RNN) was used to model dependencies amongst weather classes. \\
\end{compactenum}

\section{Proposed Deep Learning Pipeline for Near Real-time Map Generation}
\label{sec:pipeline}

In this section, we provide an overview of the pipeline for map generation using deep learning frameworks.

\subsection{Application Pipeline}

\begin{figure}[!htbp]
	\includegraphics[width=12cm]{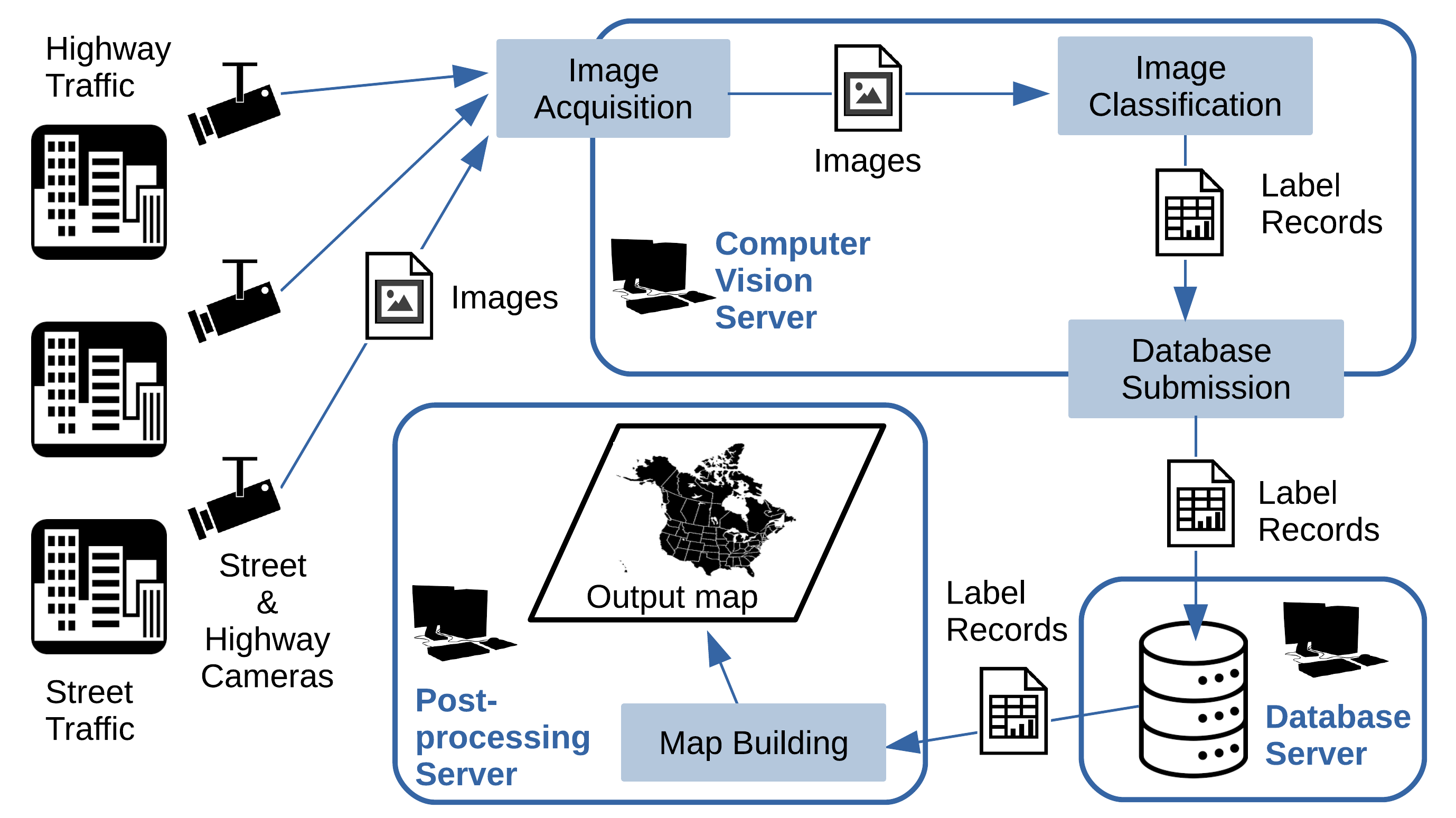}
	\centering
	\caption{A summary of near real-time road condition classification process.}
	\label{fig:pipeline}
\end{figure}

Our approach is to use CNNs to perform end-to-end classification where raw images are fed directly into the CNNs to classify road conditions. The proposed pipeline for near real-time road condition classifier consists of four modules: image acquisition, image classification, database submission and map generation. The process is summarized in Figure~\ref{fig:pipeline}. In order to reduce the overall latency, these modules are implemented in the way that they can run in an overlapping fashion. Each stage processes inputs as they emerge from the previous stage.

\begin{description}
	\item[Image Acquisition]: The first stage is the acquisition of the input images on which we perform the road condition classifications. These images are snapshots taken by street and highway cameras located across Canada and the United States. They are downloaded over the internet by sending snapshot queries to public camera APIs. For this task, we rely on a pre-assembled catalogue containing a unique camera identifier, the snapshot URL and the geographic location for each camera of interest. As the images are downloaded periodically to the computer vision server, they are passed along to the classification module for further processing. The speed of this module is practically bound to the network bandwidth and it can be executed in multiple threads.
	
	\item[Image Classification]: This is the core module performing the machine learning tasks. It contains the deep learning classifier with pre-trained weights. This module monitors the set of incoming images and are checked for corruption and integrity. If they are good, they are resized to the expected input dimensions of the classifier and fed into it in batches. The output is written into a catalogue of label records on the local disk in form of camera identifiers, time stamps, inferred label and the geographic location tags. 
	
	\item[Database Submission:] Database Submission module monitors the label records generated by the image classification module.  As they come in, they are retrieved and sent to a remote database server for further processing. This module is decoupled from the image classification task to avoid any delays.
	
	\item[Map Building]: This module  monitors the database for emerging records and fetches them to maintain an output map on which icons indicating the road conditions are super-imposed on their respective geo-locations, for visual representation.
	
\end{description}

\subsection{Operational Deep Learning Frameworks}
\label{sec:opframeworks}

In this work, we considered a variety of deep learning frameworks. The following sections give a brief overview of the various architectures used in this paper.

\subsubsection{Visual Geometry Group - VGG}

Developed by Simonyan and Zisserman~\cite{simonyan2014very}, VGG was the runner-up at the ILSVRC 2014 (ImageNet Large Scale Visual Recognition Competition). It is one of the earliest networks which showed, that using small convolutional filters with a deeply layered architecture can produce successful results. VGG has a deep feed-forward architecture with no residual connections. This is formed by linearly connected convolutional layers with max-pooling after every second or third layer, with two fully-connected layers at the end. The architecture is summarized in Table~\ref{tab:vgg-arc}.

\begin{table}[!htbp]
	\caption{\label{tab:vgg-arc}VGG Architecture Layers used in this work}
	\centering{}%
	\small
	\begin{tabular}{l|l|l}
		
		\textbf{Layer}	 	& \textbf{Kernel Size / Stride} 	& \textbf{Output Size}\\
		\hline
		Input &  & $(224 \times 224 \times 3)$ \\
		\hline
		conv-block1 $\times 2$ 	& $(3 \times 3)$ stride = 1 & ($224 \times 224 \times 64$)\\
		\hline
		maxpool1 				& $(2 \times 2)$ stride = 2 & ($112 \times 112 \times 64$)\\
		\hline
		conv-block2 $\times 2$ 	& $(3 \times 3)$ stride = 1 & ($112 \times 112 \times 128$)\\
		\hline
		maxpool2 				& $(2 \times 2)$ stride = 2 & ($56 \times 56 \times 128$)\\
		\hline
		conv-block3 $\times 3$ 	& $(3 \times 3)$ stride = 1 & ($56 \times 56 \times 256$)\\
		\hline
		maxpool3  				& $(2 \times 2)$ stride = 2 & ($28 \times 28 \times 256$)\\
		\hline
		conv-block4 $\times 2$ 	& $(3 \times 3)$ stride = 1 & ($28 \times 28 \times 512$)\\
		\hline
		maxpool4  				& $(2 \times 2)$ stride = 2 & ($14 \times 14 \times 512$)\\
		\hline
		conv-block5 $\times 2$ 	& $(3 \times 3)$ stride = 1 & ($14 \times 14 \times 512$)\\
		\hline
		maxpool5  				& $(2 \times 2)$ stride = 2 & ($7 \times 7 \times 512$)\\
		\hline
		fc-relu $\times 2$ 	& & ($4096$)\\
		\hline
		fc-softmax 	& & ($5$)\\
		\hline
		
	\end{tabular}
\end{table}

\subsubsection{Residual Neural Network- ResNet}
Developed by Kaiming He et al.~\cite{He2016}, the ResNet architecture introduces a solution to the network depth-accuracy degradation problem.  This is done by deploying shortcut connections between one or more layers of convolutional blocks that perform identity mapping, which are called residual connections. This allows the construction of a deeper network that is easier to optimize, compared to a counterpart deep network based on unreferenced mapping. ResNet won  first place in the ILSVRC 2015 classification competition. For this work, a 178-layer deep version of ResNet50 is customized for our classification experiments. The last fully connected layer is removed and replaced by a drop-out layer followed by a fully connected layer. The ResNet architecture is  given in Table~\ref{tab:resnet-arc}.

\begin{table}[!htbp]
	\caption{\label{tab:resnet-arc}ResNet Architecture Layers used in this work}
	\centering{}%
	\small
	\begin{tabular}{l|l|l}
		
		\textbf{Layer}	 	& \textbf{Kernel Size / Stride} 	& \textbf{Output Size}\\
		\hline
		Input & & $(224 \times 224 \times 3)$ \\
		\hline
		conv1 & $7 \times 7$, 64, stride 2 & ($112 \times 112 \times 64$)\\
		\hline
		maxpool & $3 \times 3$, stride 2 & ($56 \times 56 \times 64$)\\
		\hline
		conv2\textunderscore x & [$1 \times 1$, 64 $\mid$ $3 \times 3$, 64 $\mid$ $1 \times 1$, 256] $\times 3$ & ($56 \times 56 \times 256$)\\
		\hline
		conv3\textunderscore x & [$1 \times 1$, 128 $\mid$ $3 \times 3$, 128 $\mid$ $1 \times 1$, 512] $\times 4$ & ($28 \times 28 \times 512$)\\
		\hline
		conv4\textunderscore x & [$1 \times 1$, 256 $\mid$ $3 \times 3$, 256 $\mid$ $1 \times 1$, 1024] $\times 6$ & ($14 \times 14 \times 1024$)\\
		\hline
		conv5\textunderscore x & [$1 \times 1$, 512 $\mid$ $3 \times 3$, 512 $\mid$ $1 \times 1$, 2048] $\times 3$ & ($7 \times 7 \times 2048$)\\
		\hline
		global\textunderscore avg\textunderscore pooling & & 2048\\
		\hline
		dropout(rate = 0.2)& &2048\\
		\hline
		fc\textunderscore softmax&&5\\
	\end{tabular}
\end{table}

\subsubsection{InceptionResNetV2}
InceptionResNetV2 is an integration of residual connections into deep inception network \cite{szegedy2017inception}. The model achieved lower error with top-1 and top-5 error rates compared to batch normalizaion-Inception, Inception-v3, Inception-Resnet-v1 and Inception-v4. In our experiments: i) the input images were rescaled to $299 \times 299 \times 3$, ii) top layer was removed and replaced by a dropout layer with dropout rate of 0.4, and iii) with a softmax fully connected layer for the 5 classes. Details of the network are shown in Table~\ref{tab:irnv2-arc}.

\begin{table}[!htbp]
	\caption{\label{tab:irnv2-arc}InceptionResNetV2 Architecture Layers used in this work}
	\centering{}%
	\small
	\begin{tabular}{l|l|l}
		
		\textbf{Layer}	 	& \textbf{Kernel Size / Stride} 	& \textbf{Output Size}\\
		\hline
		Input & & $(299 \times 299 \times 3)$ \\
		\hline
		stem& [conv3, 32/2 V] & ($149 \times 149 \times 32$)\\ 
		& [conv3, 32 V] & ($147 \times 147 \times 32$)\\ 
		& [conv3, 64] & ($147 \times 147 \times 64$)\\ 
		& [maxpool3x3, 2 V $\|$ conv3 , 96/2 V] & ($73 \times 73 \times 160$)\\ 
		& [conv1, 64 $\mid$ conv3, 96 V $\|$ &\\
		& conv1, 64 $\mid$ conv7x1, 64 $\mid$ conv1x7, 64 $\mid$ conv3, 96 V] & ($71 \times 71 \times 192$)\\ 
		& [maxpool, 2 V $\|$ conv3 , 192 V] & ($35 \times 35 \times 256$)\\ 
		\hline
		inceptionresnet(a)$\times$ 5&5 $\times$ [conv1, 32$\|$& \\ 
		\hline
		& conv1, 32 $\mid$ conv3, 32 $\|$ & \\
		& conv1, 32 $\mid$ conv3, 48 $\mid$ conv3, 64 ] $\|$& \\
		& [conv1, 384] + ReLu &($35 \times 35 \times 256$)\\
		\hline
		reduction(a)&[ maxpool3x3, 2 V $\|$ conv3, 384 2 V $\|$ & \\
		& conv1, 256 $\mid$ conv3, l $\mid$ conv3, 384 2 V ]& ($17 \times 17 \times 896$)\\
		\hline
		inceptionresnet(b) $\times$ 10&10 $\times$ [conv1, 192 $\|$&\\
		&conv1, 128 $\mid$ conv1x7, 160 $\mid$ conv7x1, 192] $\|$ &\\
		& [conv1, 1154] + ReLu&($17 \times 17 \times 896$)\\
		\hline
		reduction(b)&[maxpool3x3, 2 V $\|$ conv1, 256 $\mid$ conv3, 384 2 V $\|$&\\
		&conv1, 256 $\mid$ conv3, 288 2 V $\|$&\\
		&conv1, 256 $\mid$ conv3, 288 $\mid$ conv3, 320 2V]&($8 \times 8 \times 1792$)\\
		\hline
		inceptionresnet(c) $\times 5$&5 $\times$ [conv1, 192$\|$&\\
		&conv1, 192 $\mid$ conv1x3, 224 $\mid$ conv3x1, 256]$\|$&\\
		&[conv1, 2048] + ReLu&($8 \times 8 \times 1792$)\\
		\hline
		avgpool&&1792\\
		\hline
		dropout(rate = 0.4)&&1792\\
		\hline
		fc\textunderscore softmax&&5\\
	\end{tabular}
\end{table}

\subsubsection{Extreme Inception - Xception}
The Xception network was introduced by Francois Chollet~\cite{xception} where Inception modules were replaced by depthwise separable convolutions with residual connections.  In the Xception architecture, the data goes through an entry flow, then a middle flow and finally an exit flow.  This process is repeated eight times. To adapt this network for our task: i) we removed the top layers and replaced them with a dropout layer, and  ii) replaced the fully connected layer with softmax for the 5 classes of road conditions.  Table ~\ref{tab:xception-arc} shows the architecture of Xception network used in this work.

\begin{table}[!htbp]
	\caption{\label{tab:xception-arc}XCeption Architecture layers used in this work}
	\centering{}%
	\small
	\begin{tabular}{l|l|l}
		
		\textbf{Layer}	 	& \textbf{Kernel Size / Stride} 	& \textbf{Output Size}\\
		\hline
		Input & & $(299 \times 299 \times 3)$ \\
		\hline
		entry&[conv3, 32/2, ReLu $\mid$ conv3, 64, ReLu] $\|$&\\
		& [conv1, 2] $\|$ [sepconv3, 128 $\mid$ReLu, sepconv3, 128 $\mid$&\\
		&maxpool3x3, 2] $\|$&\\
		&[conv1, 2] $\|$ [ReLu, sepconv3, 256 $\mid$ ReLu, sepconv3, 256 $\mid$&\\
		&maxpool3x3, 2] $\|$&\\ 
		&[conv1, 2] $\|$ [ReLu, sepconv3, 728 $\mid$ ReLu, sepconv3, 728 $\mid$&\\
		&maxpool3x3, 2] $\|$&($19 \times 19 \times 728$)\\ 
		\hline
		middle& [ReLu, sepconv3, 728 $\mid$ ReLu, sepconv3, 728 $\mid$&\\
		&ReLu, sepconv3, 728] $\|$ $\times 8$&($19 \times 19 \times 728$)\\ 
		\hline
		exit&[conv1, 2] $\|$ [ReLu, sepconv3, 728 $\mid$ ReLu, sepconv3, 1024 $\mid$&\\
		&maxpool3x3, 2] $\|$&\\
		&[sepconv3, 1536, ReLu $\mid$ sepconv3, 2048, ReLu $\mid$&\\
		&avgpool]&2048\\
		\hline
		dropout(rate = 0.2)&&2048\\
		\hline
		fc\textunderscore softmax&&5\\
		
	\end{tabular}
\end{table}

\subsubsection{EfficientNet}
EfficientNet developed by Mingxing Tan and Quoc V. Le~\cite{enet}, introduced a compound scaling method to scale up all three ConvNets dimensions, namely, width, depth and resolution, to achieve more accuracy and efficiency.  For this work, we used the baseline model EfficientNet-B0 and EfficientNet-B4. In order to apply transfer learning, we replaced the top layer with a dropout layer followed by a softmax fully-connected layer for the 5 classes. The network architecture is shown in Table~\ref{tab:efficientnet-arc}.

\begin{table}[!htbp]
	\caption{\label{tab:efficientnet-arc}EfficientNet Architecture layers used in this work}
	\centering{}%
	\small
	\begin{tabular}{l|l|l}
		
		\textbf{Layer}	 	& \textbf{Kernel Size / Stride} 	& \textbf{Output Size}\\
		\hline
		Input & & $(224 \times 224 \times 3)$ \\
		\hline
		conv3 &$3\times3$& ($224 \times 224 \times 32$)\\
		\hline
		MBconv1& $3\times3$& ($112 \times 112 \times 16$)\\
		\hline
		MBconv6 $\times 2$ & $3\times3$& ($112 \times 112 \times 24$)\\
		\hline
		MBconv6 $\times 2$ & $5\times5$& ($56 \times 56 \times 40$)\\
		\hline
		MBconv6 $\times 3$ & $3\times3$& ($28 \times 28 \times 80$)\\
		\hline
		MBconv6 $\times 3$ & $5\times5$& ($28 \times 28 \times 112$)\\
		\hline
		MBconv6 $\times 4$ & $5\times5$& ($14 \times 14 \times 192$)\\
		\hline
		MBconv6& $3\times3$& ($7 \times 7 \times 320$)\\
		\hline
		conv1&$1\times1$& ($7 \times 7 \times 1280$)\\
		\hline
		pooling&&	1280\\
		\hline
		dropout&&1280\\
		\hline
		fc\textunderscore softmax&&5\\
	\end{tabular}
\end{table}

EfficientNet-B4 is a scaled-up version of the baseline network EfficientNet-B0 by using user-specified coefficient $\phi$ and constants $\alpha$, $\beta$, $\gamma$, which are found by grid search. The latter network width, depth and resolution are determined by $\alpha^\phi$, $\beta^\phi$ and $\gamma^\phi$, respectively, under constraint of $\alpha*\beta^2*\gamma^2\approx2$.

\section{Methodology: Dataset Acquisition and Labelling} 
\label{sec:method}
A major part of this project was data acquisition, labelling and experimentation. One of main challenges in this work was to label millions of raw images of road and weather conditions variety of scenery (urban, rural), sky condition (clear, overcast), illumination (day, night, twilight) and quality to produce a reliable set of training images (shown in Figure~\ref{fig:camera}.  Another challenge  was take into account model complexity and memory usage,  in addition to classification accuracy  during the various stages of the dataset labelling and classification process. 

\begin{figure}[!htbp]
	\includegraphics[width=12cm]{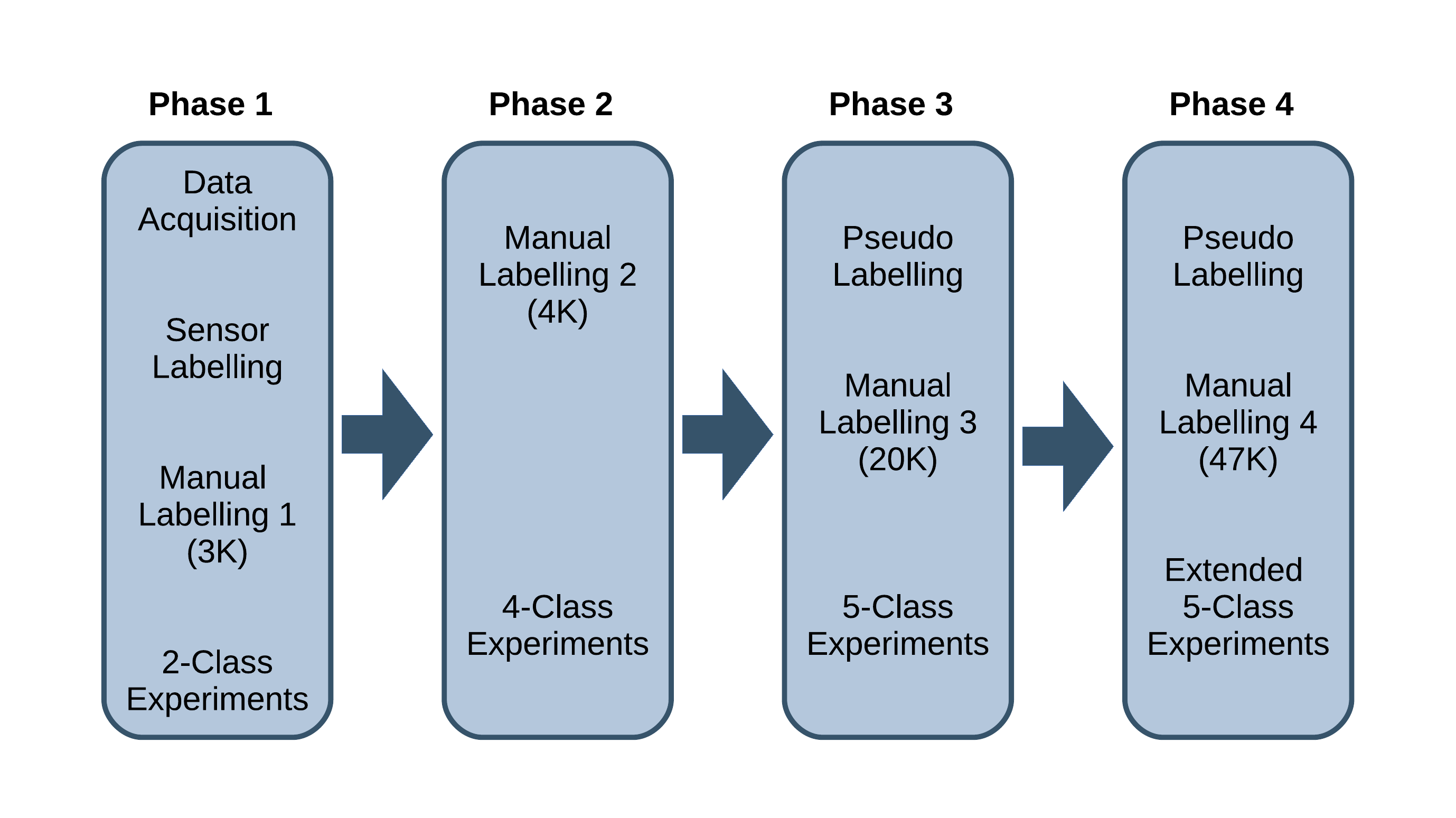}
	\centering
	\caption{An overview of the data preparation and experimentation phases.}
	\label{fig:phases}
\end{figure}

In this section, we discuss the extensive work done and explain how we proceeded in an incremental manner, from initial data collection to the final set of training examples.  The overall process is summarized in Figure ~\ref{fig:phases}.   These phases mirror some practical problems faced by the team with access to a set of live camera feeds from real-time road images collected in the month of March 2019.  The cameras span many locations across Canada and the United States depicting a wide range of road and weather conditions.  The main objective was to prepare a reliable set of labelled images for training the deep learning frameworks.   The following metrics were used in this work:

\[Precision = \frac{TP}{TP+FP}\]

\[Recall = \frac{TP}{TP+FN}\]

\[F1=2\times\frac{Precision\times Recall}{Precision+Recall}\]

\[Accuracy = \frac{TP+TN}{TP+FN+FP+TN}\]

\noindent with the usual interpretation where TP stands for True Positives, FP stands for False Positives, TN stands for  True Negatives and  FN stands for False Negatives.

\noindent The first challenge was to determine a set of (tentative) output labels for the road conditions.  Table~\ref{tab:road-condition-sensors-output} summarizes the possible classes of images inferred visually. It is noteworthy that the vast majority of the images emerged to be \emph{dry} whereas the other images spanned a variety of adverse conditions. We decided to start simple and gradually increase the number of classes. As a result, we decided to start with a binary classification task with the following classes:
  \begin{description}
	\item[Dry:] This class represented the seemingly ideal dry conditions found in a typical dry day or night images.
	
	\item[Non-dry:] This class represented non-ideal road conditions such as wet, snow, slush and ice.
\end{description}

\begin{table}[!h]
	\caption{\label{tab:road-condition-sensors-output}Labelling of images by road condition sensors located near the cameras.}
	\centering{}
		\begin{tabular}{|c|c|c|c|}
			\hline 
			\textbf{Road Condition Id} & \textbf{Title} & \textbf{\# Images} & \textbf{Content (Inferred Visually)}\\
			\hline 
			\hline 
			1 & Dry & 13,429 & Mostly dry, some wet, snow\\
			2 & Moist & 288 & Mostly damp or wet, some dry\\
			3 & Moist (treated) & 93 & Mostly damp or wet, some dry\\
			4 & Wet & 1,290 & Mostly wet, some slush,snow, dry\\
			5 & Wet (treated) & 106 & Mostly wet, some slush, snow, dry\\
			6 & Ice & 641 & Wet, some snow, slush, dry\\
			7 & Frost & 1 & Seemingly dry\\
			8 & Snow & 305 & Mostly snow, some dry, wet\\
			15 & Dew & 201 & Mostly wet, some rainy, some dry\\
			18 & Slush & 0 & N/A\\
			16 & Black ice warning & 6 & Seemingly dry\\
			21 & Unknown & 3 & Wet snowflakes on road\\
			22 & Unknown & 1 & Light snowflakes on road\\
			23 & Unknown & 108 & Mostly heavier snow or slush\\
			24 & Unknown & 213 & Seemingly assorted\\
			\hline 
		\end{tabular}
	
\end{table}

\subsection{Phase 1 - The Two-Class Problem}
\label{sec:two-class}

\subsubsection{Introduction}
In an attempt to find an alternative to manual labelling images, we tried using road condition observations from RWIS that were located near cameras. The RWIS data from departments of transportation (DOT) across North America are transmitted to the Meteorological Assimilation Data Ingest System (MADIS)\footnote{https://madis.ncep.noaa.gov/sfc\_notes.shtml\#note17}. From MADIS, we retrieved RWIS road condition observations that were located within 10 km of the camera and used them to the associated camera image. Many cameras were not close enough to an RWIS for this technique to work, but nevertheless it could reduce manual effort in numerous cases. 

In this phase, the goal was to manually prepare a two-class \emph{balanced} training set using 352,240 unlabelled images. In this process, roughly 10-20 \% of the samples were visually verifiable, so useful samples had to be cherry-picked. As can be seen from Table~\ref{tab:road-condition-sensors-output}, many of the classes were under-sampled and therefore unusable. Nonetheless, they could be accommodated together under the generic \emph{non-dry class}. By doing some cherry-picking on categories 1, 4, 5, 6, 7, 8, 16, 18, 21, 22, 23, 24 we extracted 1785 assorted non-dry samples consisting of wet, snowy, slushy and icy images. After that, we matched this class with an equal number of dry samples. Finding dry images was easy since they were abundant. Both classes span a variety of scenery (urban, rural), sky conditions (clear, overcast), illumination (day, night, twilight) and quality. We randomly split the data into training and validation sets for two classes as shown in Table~\ref{tab:3K-dataset-composition}.
\begin{table}[!htbp]
	\caption{\label{tab:3K-dataset-composition}First 2-class data set}
	\centering{}
	\begin{tabular}{|c|c|cc|}
		\hline
		& \textbf{Total} & \textbf{Train} & \textbf{Validation} \\
		\hline 
		\textbf{Dry} & 1785 & 1585 & 200 \\
		\textbf{Non-dry (Wet/Snow)} & 1785 & 1585 & 200 \\
		\hline 
		\textbf{Overall} & 3570 & 3170 & 400 \\
		\hline
	\end{tabular}
	
\end{table}

\subsubsection{Two-Class Experiments on 3K Data Set}

Once we put together our first labelled data set, we considered two DCNN architectures for our first round of experiments.

\begin{description}
	\item[VGG-16:] This architecture was chosen for a number of reasons: i) demonstrably successful on a variety of image classification tasks, ii) has a feed-forward architecture with no residual connections which makes it a good baseline, iii) has native support in Keras\footnote{https://keras.io/} and its model with weights are publicly available, and iv)  was computationally feasible. VGG has different flavours but a popular one, which has 13 convolutional and pooling layers and 3 fully connected layers, is called VGG-16. We used its Tensorflow implementation with Keras. We took the original VGG-16 classifier with 1000 classes, discarded the final fully-connected layer and appended a new one with 2 neurons activated via the softmax function. We used the default ImageNet weights. We set every layer but the final one non-trainable. It's second-to-last layer is fully connected with 4096 neurons which means we ended up with 4096 x 2 (weights) + 2 (bias) = 8194 trainable parameters. All the previous 134,260,544 parameters were left frozen.
	
	\item[ResNet-50:] This architecture uses residual connections to tie non-adjacent layers with the intent of coping with vanishing/exploding gradients during the training process. We started with ResNet-50, a fifty layer deep version of this architecture. As with VGG, we used Keras with Tensorflow. We took the original classifier and configured the end layers for our 2-class problem. The final layer had 2 neurons activated via the softmax function. It's second-to-last layer was fully connected with 2048 neurons so there were 2048 x 2 (weights) + 2 (bias) = 4098 trainable parameters. All previous layers with 23,587,712 parameters were set to non-trainable.

\end{description}

\noindent For both VGG-16 and ResNet50, we used a batch size of 10 and we trained both the networks for 20 epochs with increments of five using our 3170 training samples. Then we tested the models on our 400 validation samples. The classification report and the confusion matrix are presented in Tables~\ref{tab:phase1-classrep-vgg-resnet} and~\ref{tab:phase1-confmat-vgg-resnet}.

\begin{table}[!htbp]
	\caption{\label{tab:phase1-classrep-vgg-resnet}Classification reports for VGG-16 and ResNet50 after epoch 20.}
	\centering{}
		\begin{tabular}{|c|cccc|c|}
			\hline 
			\textbf{ResNet-50} & \textbf{Precision} & \textbf{Recall} & \textbf{F1-Score} & \textbf{Support} & \textbf{Accuracy}\\
			\hline 
			\textbf{Dry} & 0.61 & 0.96 & 0.75 & 200 & \\
			\textbf{Non-dry} & 0.91 & 0.38 & 0.53 & 200 & \\
			\hline 
			\textbf{Training} &  &  &  &  & 88.7\%\\
			\textbf{Validation} &  &  &  &  & \textbf{67.0\%}\\
			\hline 
			\hline
			\textbf{VGG-16} & \textbf{Precision} & \textbf{Recall} & \textbf{F1-Score} & \textbf{Support} & \textbf{Accuracy}\\
			\hline 
			\textbf{Dry} & 0.79 & 0.88 & 0.83 & 200 & \\
			\textbf{Non-dry} & 0.86 & 0.77 & 0.81 & 200 & \\
			\hline
			\textbf{Training} &  &  &  &  & 94.0\%\\
			\textbf{Validation} &  &  &  &  & \textbf{82.3\%}\\
			\hline 
		\end{tabular}
	
\end{table}

\begin{table}[!htbp]
	\caption{\label{tab:phase1-confmat-vgg-resnet}Confusion matrices showing true labels vs. predicted labels after epoch 20.}
	\centering{}
	\begin{tabular}{|c|cc|cc|}
		\hline 
		& \multicolumn{2}{c|}{\textbf{ResNet-50}} & \multicolumn{2}{c|}{\textbf{VGG-16}}\\
		\hline 
		\textbf{T \textbackslash{} P} & \textbf{Dry} & \textbf{Non-dry} & \textbf{Dry} & \textbf{Non-dry}\\
		\hline 
		\textbf{Dry} & \textbf{193} & 7 & \textbf{176} & 24\\
		\textbf{Non-dry} & 125 & \textbf{75} & 47 & \textbf{153}\\
		\hline 
	\end{tabular}
		
\end{table}

\subsubsection{Analysis}

Based on our experiments in Phase 1, it can be seen that the classifiers have been able to differentiate between dry and non-dry images, showing promise for the upcoming multi-label classification tasks. Also, VGG-16 outperformed ResNet-50 in terms of overall accuracy and F1 scores. Perhaps unsurprisingly, both frameworks had less difficulty classifying the dry class since it had higher intra-class similarity. On the contrary, the non-dry class spanned a wider range of conditions resulting in a higher intra-class variation.

The main importance of these results is that they showed we can employ modern DCNN frameworks along with transfer learning to achieve non-trivial results. Another interesting thing to note is that these results were achieved without performing any pre-processing on the images (noise removal, text/logo removal, histogram equalization, cropping etc.) other than rescaling. This suggests that modern architectures have the potential to adapt to our scene classification problem. We should also note that throughout the first round of data labelling and experimentation, we identified a number of potential challenges for interpreting the image content. They include:

\begin{description}
	\item[Resolution:] Images had varying resolutions and aspect ratios. The majority were 320 by 240 but they ranged from 160 by 90 to 2048 by 1536. Modern CNN architectures expect inputs of uniform size so they would be required to resized accordingly. It does, however, mean that the solution we develop would need to be scale-invariant.
	
	\item[Illumunation:] Depending on the time of the day, some images yielded very dark scenes, making it practically impossible to judge the road condition. 
	
	\item[Corruption:] We came across some corrupted images containing regions with pixelation and unnatural colours.
	
	\item[Occlusion:] Certain images had objects partly or fully blocking the view. 
	
	\item[Superimposed Texts:] Most images contained super-imposed logo and text on the camera view. These would have to be sampled adequately across the classes to prevent our model to use them as features.
	
	\item[Varying Angles:] The angle at which the cameras view the road varies greatly. Some cameras have top-down views, others are almost at eye-level.  
	
	\item[Varying Distance:] The distance between the cameras and the roads also vary a lot. 
	
	\item[Imbalanced Categories:] Vast majority of the images were the ideal dry condition. Conditions like snow, slush and ice seemed significantly less frequent.
	
	\item[Offline Cameras:] From time to time, cameras show ``stream offline" message rather than the actual video feed.
	 
\end{description}

Based on the promising results we achieved with VGG-16 on a simplified two-class configuration, we decided to use the VGG-16 architecture to expand the task to a multi-label (4-class) classification problem.

\subsection{Phase 2 - The Four-Class Problem}
\label{sec:four-class}

\subsubsection{Introduction}
The natural way to convert this into a multi-label problem was to decompose the non-dry class into its constituents.  However, the subclasses were very imbalanced in terms of size and some of them had negligible useful content. The subclasses were categorized under two streams of roughly equal sizes: ``wet" and ``snow". Accordingly, we split the non-dry class with 1785 images into roughly equal wet and snow classes.

In addition, we also observed another particularly interesting group of images. There were thousands of snapshots from \emph{offline} cameras. These belonged to the cameras for which the video stream from the view sites were not available so the camera sent its own ``stream is offline" image. These images would also be present in the production environment and it was important to capture and filter such images.   As a result, 677 assorted offline images were extracted and the classification task in this phase used the following classes:

\begin{description}
	\item[Dry:] This class represented the seemingly ideal dry condition yielded by a typical dry day or night images.
	
	\item[Wet:] This class represented a spectrum of conditions from moist roads to puddles to soaking wet.
	
	\item[Snow:] This class represented harsh winter conditions including snow-covered, slush-covered and ice-covered roads.
	
	\item[Offline:] This class covered the static no-signal feed of the cameras which varied from camera to camera.
\end{description}

\noindent Table~\ref{tab:4K-dataset-composition} shows the new data set composition. Note, that at this stage, we were also using different sample distributions per class than the ones used in Section \ref{sec:two-class} with mostly dry images. This is because we also wanted to observe the behaviour of the classifier over an unevenly distributed data set, reflecting the composition of the 352K data set, which in turn reflected the underlying weather conditions across the continent.  This was meant to mirror the conditions encountered in a  real-time production environment.

\begin{table}[!htbp]
	\caption{\label{tab:4K-dataset-composition}First 4-class data set}
	\centering{}
	\begin{tabular}{|c|c|cc|}
		\hline
		& \textbf{Total} & \textbf{Train} & \textbf{Validation} \\
		\hline 
		\textbf{Dry} 		& 1785 & 1696 & 89 \\
		\textbf{Offline} 	& 677  & 644  & 33 \\
		\textbf{Snow} 		& 905  & 860  & 45 \\
		\textbf{Wet} 		& 880  & 837  & 43 \\
		\hline 
		\textbf{Overall} 	& 4247 & 4035 & 210 \\
		\hline
	\end{tabular}
		
\end{table}

\subsubsection{Four-Class Experiments on 4K Data Set}

For this stage, we decided to repurpose the VGG-16 classifier from Section \ref{sec:two-class} since it showed more promise. We essentially employed the same hyper-parameters as Section \ref{sec:two-class} except that we changed the final layer to include 4 neurons for our four-class setup. We split the labelled data as 95\% training and 5\% validation. We reserved a smaller portion for validation since we had more classes with less images. After training a VGG-16 classifier for 5 epochs, the training set accuracy was 83.5\%  and the validation set accuracy was 77.1\%. The classification report and the confusion matrix for the validation set are presented in Tables~\ref{tab:phase2-vgg-ssl-classrep} and~\ref{tab:phase2-vgg-ssl-confmat} respectively. 

\begin{table}[!htbp]
	\caption{\label{tab:phase2-vgg-ssl-classrep}Classification report for validation set
		after epoch 5.}
	\centering{}
	\begin{tabular}{|c|cccc|c|}
		\hline 
		\textbf{VGG-16} & \textbf{Precision} & \textbf{Recall} & \textbf{F1-Score} & \textbf{Support} & \textbf{Accuracy}\\
		\hline 
		\textbf{Dry} & 0.81 & 0.78 & 0.79 & 89 & \\
		\textbf{Offline} & 0.97 & 1.00 & 0.99 & 33 & \\
		\textbf{Snow} & 0.83 & 0.78 & 0.80 & 45 & \\
		\textbf{Wet} & 0.51 & 0.58 & 0.54 & 43 & \\
		\hline 
		\textbf{Training} &  &  &  &  & 83.5\%\\
		\textbf{Validation} &  &  &  &  & \textbf{77.1\%}\\
		\hline 
	\end{tabular}

\end{table}

\begin{table}[!htbp]
	\caption{\label{tab:phase2-vgg-ssl-confmat}Confusion matrix showing true labels
		vs predicted labels for validation set after epoch 5.}
	\centering{}
	\begin{tabular}{|c|cccc|}
		\hline 
		\textbf{T \textbackslash{} P} & \textbf{Dry} & \textbf{Offline} & \textbf{Snow} & \textbf{Wet}\\
		\hline 
		\textbf{Dry} & \textbf{69} & 0 & 2 & 18\\
		\textbf{Offline} & 0 & \textbf{33} & 0 & 0\\
		\textbf{Snow} & 3 & 1 & \textbf{35} & 6\\
		\textbf{Wet} & 13 & 0 & 5 & \textbf{25}\\
		\hline 
	\end{tabular}

\end{table}

\subsubsection{Analysis}

This round of experimentation resulted in a mixed set of results. The overall accuracy decreased to 77\%, although this was still promising since we had twice as many labels and the decomposed wet and snow classes had effectively half as many samples to work with. A detailed class-based analysis is presented below:

\begin{description}
	\item[Dry:] Dry performed slightly worse than the two-class experiment, mainly due to the fact that the model had difficulty distinguishing between the dry and the wet classes. These classes seemed to have relatively higher inter-class similarity. More training samples were necessary.
	
	\item[Wet:] This class suffered more than the dry class mainly because of its smaller size. We think when these samples were represented together with the snow class, they were easier to tell apart from the dry images because wet and snow actually have many common features such as less visibility and more frequent overcast scenes. These features did not appear to be emphasized as much by the model since they are now associated with multiple classes.
	
	\item[Snow:] This class offered similar results to the two-class experiments, which is impressive considering the fact that it had effectively half as many examples as in Section \ref{sec:two-class}. This is most likely because the snow class had more distinctive features compared to the dry and wet class, such as the colour and the texture of the road being much different.
	
	\item[Offline:] The classifier performed exceptionally well (with 100\% accuracy) over the new offline category. We think this is because of its distinctive features such as large texts and unnatural colours, which are totally different from a regular road scene.
\end{description}

Overall, the experiments showed that VGG-16 was still able to show a decent performance on a four-class setup, even with a small subset of images that had an  uneven training data distribution. Validation results suggest that the highest degree of confusion was between dry and wet categories, whereas the number of mistakes among the other pairs were smaller. Remarkably, the model was exceptionally good at classifying offline images. In the next phase, we explored the ways to increase the size of our labelled data set.

\subsection{Phase 3 - The Five-Class Problem}
\label{sec:five-class}

\subsubsection{Pseudo-Labelling with VGG}

In Phase 3, the goal  was to leverage the VGG-16 classifier trained on the four- class experiments and use a semi-supervised learning method on the large 352K data set to ``pseudo-label" each image with one of our four classes.  This approach was used to assist us in clustering images of similar nature, making the manual cherry-picking stage much easier.  We incorporated our 4-label model from Section~\ref{sec:four-class} to classify the 352K images  which took under a day on an Intel i7 3630QM machine with 16GB system RAM and NVIDIA GeForce 670MX GPU with 3GB video RAM. Table~\ref{tab:image-set1} shows the number of images suggested per class by VGG-16, along with the training and validation data that we provided.

\begin{table}[!htbp]
	\caption{\label{tab:image-set1}Number of images VGG-16 pseudo-labelled per class, as suggestions.}
	\centering{}
	\begin{tabular}{|c|ccc|c|}
		\hline 
		& \multicolumn{3}{c|}{\textbf{Training \& Validation}} & \textbf{Classification}\\
		\textbf{\# Images} & \textbf{Train} & \textbf{Valid} & \textbf{Total} & \textbf{VGG-16 Suggestions}\\
		\hline 
		\textbf{Dry} 		& 1696 	& 89 	& 1785 	& 142,413\\
		\textbf{Wet/Moist} 	& 837 	& 43 	& 880 	& 99,089\\
		\textbf{Snow/Slush} & 860 	& 45 	& 905 	& 70,200\\
		\textbf{Offline} 	& 644 	& 33 	& 677 	& 40,538\\
		\hline 
		\textbf{Overall} 	& 4035 	& 210 	& 4247 	& 352,240\\
		\hline 
	\end{tabular}

\end{table}

At this stage, we did not use any metrics for the classification accuracy over the 352K set as they were mostly unlabelled data. However, on examining the  thumbnails, we observed similar types of classification errors as in previous experiments on the validation set, in the sense that dry and wet samples were being identified incorrectly on dark and overcast imagery, dry winter samples were sometimes getting mistaken for snow, and we were seeing exceptionally good results for the offline class. An overview of the extracted content by means of semi-supervised learning is presented below:

\begin{description}
	\item [Dry:] Consisted of mostly clear and well-lit scenes with dry and
	clean pavement. Day and summer-looking images were frequent whereas
	winter and night images were sparse. There were occasional images with rain
	or wet/moist pavement, a small number of snowy background images or very
	sparse snowy road images, a  negligible number of offline images. Precision
	was seemingly good, but recall was not, since dirty, overcast, night and
	dark pavement images leaked to the wet class and winter images leaked
	to the snow class.
	
	\item [Wet:] Included a lot of rainy and wet surfaces. In addition, significant
	portions of the images were overcast dry, fuzzy, foggy, blurry, dark and night scenery,
	dirty or otherwise unintelligible scenes (not necessarily wet).
	Infrequent snow and wet-snow images were present. There were a negligible number of offline images.
  Neither recall nor precision were good, but nevertheless there were sufficient wet scenes to
	harvest for our purposes.
	
	\item [Snow:] Recall was good, seemingly covering the majority of snowy
	road scenes across all categories. However, precision was poor. Many wet and dry images appeared under this class, although most
	of them were winter scenes with snowy background, but clear road. Also,
	the model confused a certain shade of road pavement and texture with snow.
	As with the ``wet'' category, we observed fuzzy, blurry, foggy, dark and night
	scenery, dirty objective or otherwise illegible scenes (not necessarily
	snowy). None or negligible offline imagery were present.
	
	\item [Offline:] This was by far the most accurate class. Recall was near-perfect and
	precision was good. It encapsulated almost all offline imagery
	along with some almost-black pictures. It learned the offline
	patterns that it had not been not trained. We saw a very small percentage of
	other kinds of errors. 
		
\end{description}

\subsubsection{Manual Labelling}

Training a 4-class VGG-16 network and running it on the entire 352K data set helped us cluster similar images and significantly narrow down sets of potentially interesting images for each class.  We were able to manually label approximately 9K dry, 5K wet and 4K snow samples.  During this phase, we observed that a huge portion of poorly-lit night images ended up in either snow or wet categories.  These images (1108) were very hard to label even by human classifiers and hence led to a new category ``poor''.   The total number of images resulting from this phase was approximately 20K.    Table~\ref{tab:20K-dataset-composition} shows the composition of our 5-class data set with 20K images. The description of the fifth class is as follows:
 
\begin{description}
	\item[{Poor}]: This class contains images that a human was unable to classify, due to various factors including: darkness, poor visibility, blurriness, or uncertainty about the category (e.g. wet vs. ice). 
\end{description}

\begin{table}[!htbp]
	\caption{\label{tab:20K-dataset-composition}First 5-Class Data Set.}
	\centering{}
	\begin{tabular}{|r|c|cc|}
		\hline 
		& \textbf{Total} & \textbf{Train} & \textbf{Validation}\\
		\hline 
		\textbf{Dry} & 9620 & 7696 & 1924\\
		\textbf{Wet} & 5012 & 4009 & 1003\\
		\textbf{Snow} & 4028 & 3222 & 806\\
		\textbf{Offline} & 676 & 540 & 136\\
		\textbf{Poor/Dark} & 1108 & 886 & 222\\
		\hline 
		\textbf{Overall} & 20,444 & 16,353 & 4091\\
		\hline 
	\end{tabular}
	
\end{table}

\subsubsection{Five-Class Experiments on 20K Data Set}
\begin{figure}[!htbp]
	\centering{}
		\includegraphics[width=0.2\textwidth]{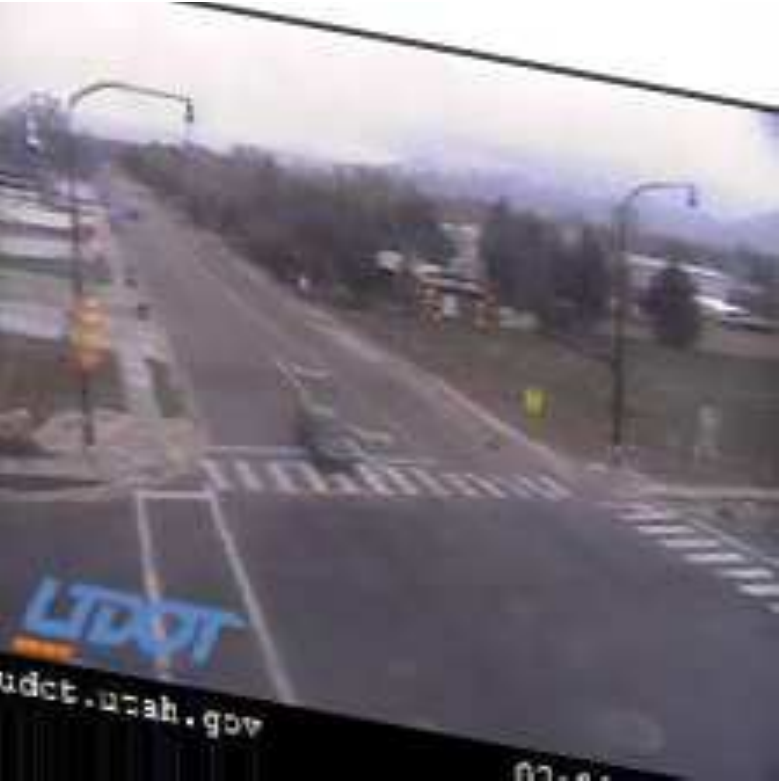}\,\includegraphics[width=0.2\textwidth]{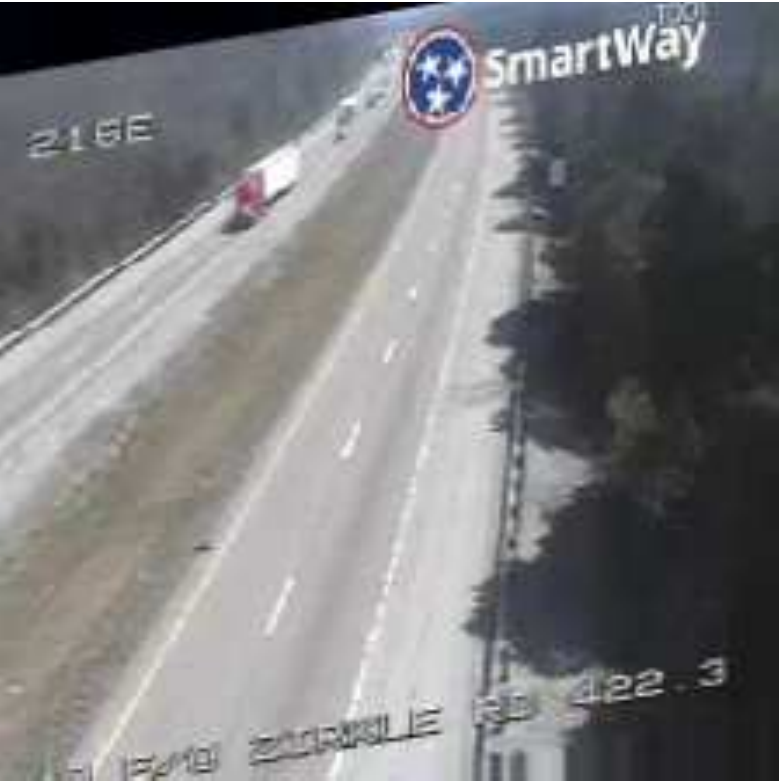}\,\includegraphics[width=0.2\textwidth]{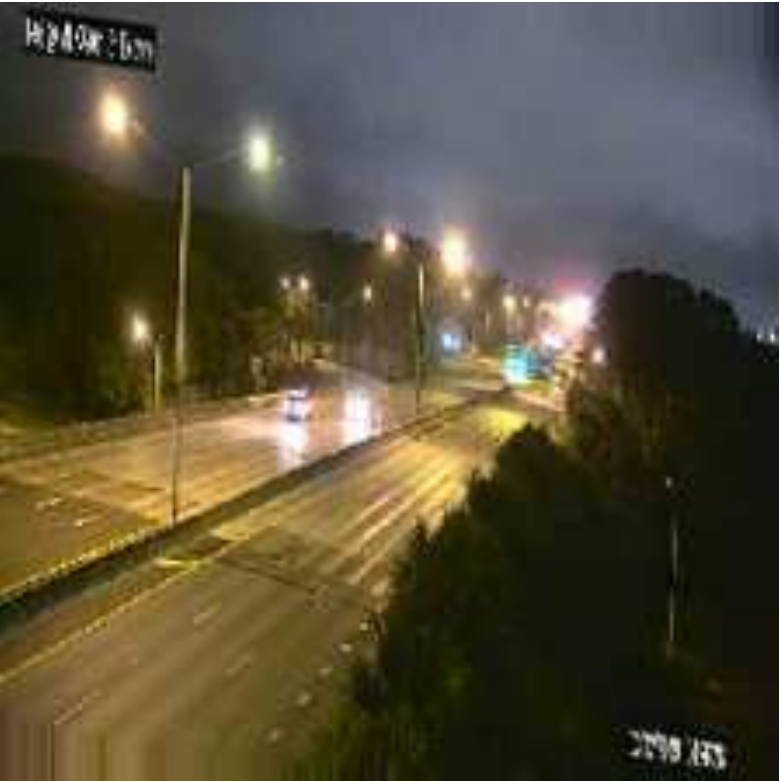}\,\includegraphics[width=0.2\textwidth]{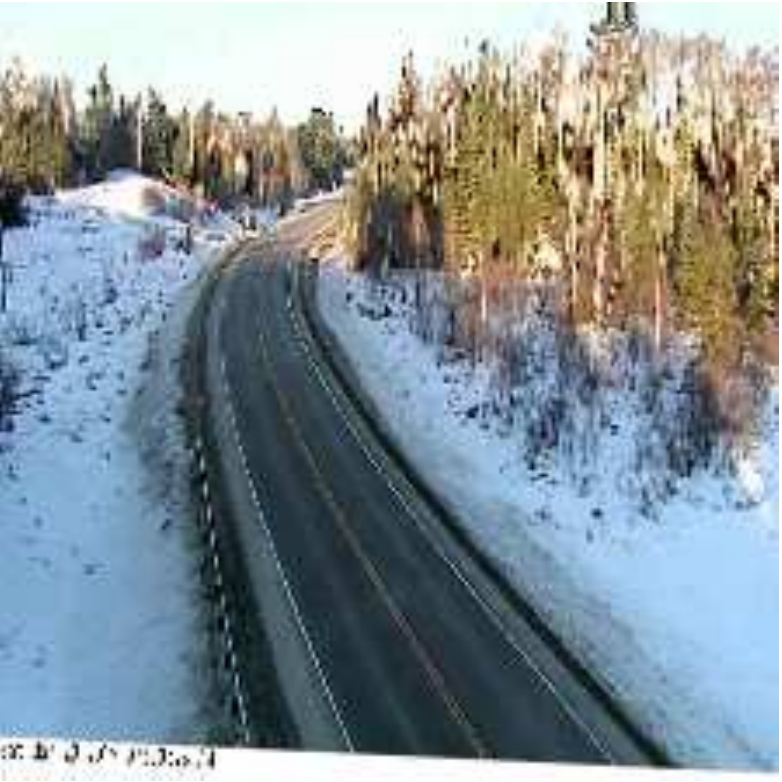}
		
	\caption{\label{fig:Augmented-data-samples}Augmented data samples}
\end{figure}

For this round of experimentation, we decided to retrain the VGG-16 classifier for two reasons. Firstly, because we had more substantial amount of samples for training and we wanted to allow more layers to be tuned. In order to increase the fitting capacity of the model, we planned to train all the fully connected layers along with the last few convolutional layers. However, we came across over a hundred million parameters to be tuned, which our hardware could not handle. Therefore, we needed to reduce the number of parameters by shrinking the fully connected layers. The standard form of VGG-16 contained 4096 neurons per hidden layer which was reduced to 1024. Secondly, we realized we could further diversify our data set and hopefully improve the performance with data augmentation. This technique introduces additional training samples to the data set by transforming and deforming the originals (randomly shifting, rotating, zooming, flipping etc.) as shown in Figure \ref{fig:Augmented-data-samples}. Although new samples would be correlated with the original images, they could counteract over-fitting in some cases as the transformations and deformations challenge the training process.  Table~\ref{tab:exp-config} summarizes the experimental setup.

\begin{table}[!htbp]
	\caption{\label{tab:exp-config}Experimental configuration for VGG-16.}
	\centering{}
	\begin{tabular}{|ll|}
		\hline 
		\textbf{Hyper-parameter} & \textbf{Value}\\
		\hline 
		Number of classes & : 5\\
		Number of epochs & : 10\\
		Batch size & : 16\\
		Optimizer algorithm & : Adam\\
		Learning rate & : 0.0001\\
		Validation split & : 0.2\\
		Base architecture & : VGG\\
		Convolutional layers & : Default VGG-16\\
		Trainable conv. layers & : Last 4 (Conv. $>$  Conv. $>$ Conv. $>$ Max-Pool)\\
		Fully connected layers & : ReLU-1024 $>$ Dropout-50\% $>$ Softmax-3\\
		Trainable FC layers & : All\\
		Total parameters & : 48,474,693\\
		Trainable parameters & : 40,839,429\\
		Non-trainable parameters & : 7,635,264\\
		Parameter initialization & : Default VGG-16 (ImageNet)\\
		\hline 
		\textbf{Data Augmentation} & \textbf{}\\
		\hline 
		Width shift range & : 0.1\\
		Height shift range & : 0.1\\
		Shear range & : 0.01\\
		Zoom range & : {[}0,9, 1.0{]}\\
		Horizontal flip & : True\\
		Vertical flip & : False\\
		Fill mode & : Constant\\
		Brightness range & : {[}0.5, 1.5{]}\\
		\hline 
	\end{tabular}

\end{table}

We used 80-20 split and trained this VGG-16 classifier initialized with ImageNet\footnote{http://www.image-net.org/} weights. The results are presented in Tables~\ref{tab:phase3-vgg-classrep} and~\ref{tab:phase3-vgg-confmat}.

\begin{table}[!htbp]
	\caption{\label{tab:phase3-vgg-classrep}Classification report for validation set
		after epoch 5.}
	\centering{}
	\begin{tabular}{|c|cccc|c|}
		\hline 
		\textbf{VGG-16} & \textbf{Precision} & \textbf{Recall} & \textbf{F1-Score} & \textbf{Support} & \textbf{Accuracy}\\
		\hline 
		\textbf{Dry} 		& 0.88 & 0.90 & 0.93 & 1924 & \\
		\textbf{Offline} 	& 1.00 & 0.99 & 0.98 & 136 &\\
		\textbf{Poor} 		& 0.91 & 0.91 & 0.91 & 222 &\\
		\textbf{Snow} 		& 0.92 & 0.91 & 0.91 & 806 &\\
		\textbf{Wet} 		& 0.87 & 0.83 & 0.79 & 1003 &\\
		\hline 
		\textbf{Training} 	&  &  &  &  & 94.1\%\\
		\textbf{Validation} &  &  &  &  & \textbf{89.0\%}\\
		\hline 
	\end{tabular}

\end{table}

\begin{table}[!htbp]
	\caption{\label{tab:phase3-vgg-confmat}Confusion matrix showing true labels
		vs predicted labels for validation set after epoch 5.}
	\centering{}
		\begin{tabular}{|c|ccccc|}
			\hline 
			\textbf{T \textbackslash{} P} & \textbf{Dry} & \textbf{Offline} & \textbf{Poor} & \textbf{Snow} & \textbf{Wet}\\
			\hline 
			\textbf{Dry} 		& \textbf{1780} 	& 0 	& 6 	& 41 	& 97 \\
			\textbf{Offline} 	& 0 	& \textbf{133} 	& 3 	& 0 	& 0 \\
			\textbf{Poor} 		& 9 	& 0 	& \textbf{202} 	& 5 	& 6 \\
			\textbf{Snow} 		& 55 	& 0 	& 1 	& \textbf{733} 	& 17 \\
			\textbf{Wet} 		& 181 	& 0 	& 9 	& 19 	& \textbf{794} \\
			\hline 
		\end{tabular}
	
\end{table}

\subsubsection{Analysis}

Experiments in this phase yielded the best results yet. The F-1 scores for dry, wet, and snow all increased as we quadrupled our labelled data set and incorporated data augmentation. The new ``poor" class also performed well, achieving an F-1 score of 91\%. The most challenging task was still distinguishing dry and wet images, although it should be noted that we saw a big improvement in the F-1 score for the wet class from 54\% to 79\% with the additional samples.

Another point worth mentioning is that at the end of this phase, we achieved the lowest deviation so far, between the peak validation accuracy and the underlying training accuracy. This suggests that over-fitting became less of a concern at this point and we got a better representation of the images in our model, extending its performance on the training set to the validation set relatively well.  It should be observed that this result was obtained with a cherry-picked manually labelled 20K dataset.  During this time, we also had the chance to acquire \emph{additional} road camera samples beyond our 352K data set.
These samples were collected using the same method described in Section \ref{sec:two-class} by randomly sampling cameras across the North America at different times of the day, over a period of three months in all kinds of climate and weather conditions. Combined with our earlier unlabelled images, we were able to generate a 1.5 million-image data set.  

We extracted 1000 images randomly and we let our VGG-16 model classify this set. Then we manually interpreted the results. Taking into account the ambiguous and fuzzy cases which could belong to multiple labels, we marked each result as ``acceptable" which were most likely (or at least partially) correct or ``refused" which were absolute false positives. Table~\ref{tab:random1000-20K} shows these results where the overall classification accuracy was good (88\%) for the dry, offline and poor classes. However, the performance on wet and snow categories was unacceptable. There were too many false-positives.  Upon further examination, we realized most of them were poor images incorrectly assigned to these classes.  Therefore, we decided to diversify our training set with more low-quality samples.

\begin{table}[!htbp]
	\caption{\label{tab:random1000-20K}Classification judgment for 1000 random images
		from combined (1.5M) data sets}
	
	\centering{}%
	\begin{tabular}{|c|cccccc|c|}
		\hline 
		& \textbf{Verdict} & \textbf{Dry} & \textbf{Offline} & \textbf{Poor} & \textbf{Snow} & \textbf{Wet} & \textbf{Total}\\
		\hline 
		\textbf{VGG-16} & Acceptable: & 616 & 65 & 136 & 9 & 62 & 888\\
		\textbf{20K} & \textbf{Refused:} & \textbf{31} & \textbf{0} & \textbf{7} & \textbf{21} & \textbf{53} & \textbf{112}\\
		\hline 
	\end{tabular}
\end{table}
 
\subsection{Phase 4 - Scaling the Solution}
\label{sec:five-class-diverse}

\subsubsection{Second Pseudo-Labelling}

In this phase,  we decided to perform another round of labelling using semi-supervised learning as in Section~\ref{sec:four-class}. This time, we applied our five-label VGG-16 classifier trained in Section~\ref{sec:five-class} with 89\% validation accuracy on both 352K data set and the 1.1M data set. Table ~\ref{tab:B1-B2-Compared} shows the number of images pseudo-labelled by class.

\begin{table}[!htbp]
	\caption{\label{tab:B1-B2-Compared}VGG-16 pseudo-labelling results the 352K and 1.1M data sets.}
	\centering{}
	\begin{tabular}{|c|rr|rr|}
		\hline 
		\textbf{VGG-89\%} & \multicolumn{2}{c|}{\textbf{352K Set}} & \multicolumn{2}{c|}{\textbf{1.1M Set}}\\
		\textbf{Pseudo-Labels} & \textbf{\# Images} & \textbf{\%} & \textbf{\# Images} & \textbf{\%}\\
		\hline 
		\textbf{Dry} & 216,614 & 61.50 & 748,165 & 67.16\\
		\textbf{Poor} & 49,707 & 14.11 & 143,016 & 12.84\\
		\textbf{Wet} & 43,479 & 12.34 & 114,700 & 10.30\\
		\textbf{Offline} & 26,827 & 7.62 & 91,163 & 8.18\\
		\textbf{Snow} & 15,613 & 4.43 & 16,876 & 1.52\\
		
		\hline 
		\textbf{Total} & \textbf{352,240} & \textbf{100.00} & \textbf{1,113,920} & \textbf{100.00}\\
		\hline 
	\end{tabular}
		
\end{table}
\noindent For the 352K data set, the number of dry samples increased whereas the others decreased (compared to the results in Table~\ref{tab:image-set1}). This seemed to be a step in the right direction, since many of the poor samples already migrated to the poor class after it was introduced. Also, in Section ~\ref{sec:four-class} those classes were suffering from low precision, so having fewer samples appear in those categories was a good sign, suggesting less false positives and a higher precision. In the 1.1M data set, we had a similar distribution, albeit not the same. Here, the dry class samples dominated even more. Also, snow samples were especially rare since those images were captured later in spring.

\subsubsection{Manual-Labelling}

In our earlier process of generating labelled sets, we used suggestions by the VGG-16 network to cherry-pick high confidence samples and ignored the remaining majority samples which resulted in a high validation accuracy. However, we were still facing many false positives and ambiguous samples. So in this phase,  we decided to have all types of images represented in our model for training and validation.

Accordingly, we considered the 1.5 million images with their pseudo-labels and we randomly extracted 4000 samples from each of the five classes. Then we manually labelled them. This time, unlike in Section~\ref{sec:four-class}, we did not cherry-pick the results. We did not discard any image, but included them in the ``poor" class if we were unable to label them with confidence. So now, the poor class represented all types of problematic images including the \emph{challenging} images discussed in Section~\ref{sec:two-class}. 

To ensure that we did not trivialize our original high confidence  cherry-picked training samples by the new randomly extracted images, we also performed another set of cherry-picking, although to a lesser degree. Overall, we supplemented our 20K data set with 20K randomized and 7K cherry-picked samples. Table \ref{tab:47K-dataset-composition} shows the final labelled data set.

\begin{table}[!htbp]
	\caption{\label{tab:47K-dataset-composition}Final 47K labelled data set with 5 classes.}
	\centering{}
	\begin{tabular}{|r|c|cc|}
		\hline 
		& \textbf{Total} & \textbf{Train} & \textbf{Validation}\\
		\hline 
		\textbf{Dry} 		& 16,065 & 14,458 & 1607\\
		\textbf{Offline} 	& 5225 	& 4702 & 523\\
		\textbf{Poor} 		& 9259 	& 8333 & 926\\
		\textbf{Snow} 		& 7565 	& 6808 & 757\\
		\textbf{Wet} 		& 9228 	& 8305 & 923\\
		\hline 
		\textbf{Overall} 	& 47,342 & 42,606 & 4736\\
		\hline 
	\end{tabular}
		
\end{table}

\subsubsection{Five-Class Experiments on 47K Data Set}

At this stage of the project, we upgraded our development machine to a Intel i7 9700K CPU, 32GB RAM and NVIDIA GeForce RTX 2080 with 8GB Video RAM. This enabled us to experiment with more advanced frameworks such as InceptionResNetV2 and EfficientNet-B4 in addition to VGG-16. InceptionResNetV2 is a very deep and sophisticated architecture with 55 million parameters spread across 572 layers compared to the VGG-16 model with 48 million parameters over 23 layers (including non-convolutional layers). Introduced recently by Google AI, EfficientNet leverages scalability in multiple dimensions, unlike most other networks. It employs 18M parameters. For all three frameworks, we used our 47K data set via 90-10 training/validation split and data augmentation to maximize the training samples. As we multiplied the size of our data set, this ratio yielded a similar number of validation examples as Section~\ref{sec:five-class}. We now give the results of experiments with the three frameworks.

\begin{description}
	\item [VGG-16:] It was trained with our previous hyper-parameters except that we used 1280 neurons in the fully connected layers, which resulted in 40 million parameters to be trained. Table ~\ref{tab:phase4-vgg-classrep} shows the classification report and Table~\ref{tab:phase4-vgg-confmat} shows the confusion matrix after epoch 7. 
\end{description}

\begin{table}[!htbp]
	\caption{\label{tab:phase4-vgg-classrep}Classification report for validation set
		after epoch 7.}
	\centering{}
	\begin{tabular}{|c|cccc|c|}
		\hline 
		\textbf{VGG-16} & \textbf{Precision} & \textbf{Recall} & \textbf{F1-Score} & \textbf{Support} & \textbf{Accuracy}\\
		\hline 
		\textbf{Dry} 		& 0.87 & 0.88 & 0.87 & 1607 & \\
		\textbf{Offline} 	& 0.99 & 0.98 & 0.99 & 523 &\\
		\textbf{Poor} 		& 0.88 & 0.84 & 0.86 & 926 &\\
		\textbf{Snow} 		& 0.90 & 0.85 & 0.88 & 757 &\\
		\textbf{Wet} 		& 0.80 & 0.85 & 0.82 & 923 &\\
		\hline 
		\textbf{Training} 	&  &  &  &  & 88.9\%\\
		\textbf{Validation} &  &  &  &  & \textbf{87.3\%}\\
		\hline 
	\end{tabular}
	
\end{table}

\begin{table}[!htbp]
	\caption{\label{tab:phase4-vgg-confmat}Confusion matrix for VGG-16 showing true labels
		vs predicted labels for validation set after epoch 7.}
	\centering{}
	\begin{tabular}{|c|ccccc|}
		\hline 
		\textbf{T \textbackslash{} P} & \textbf{Dry} & \textbf{Offline} & \textbf{Poor} & \textbf{Snow} & \textbf{Wet}\\
		\hline 
		\textbf{Dry} 		& \textbf{1419} 	& 0 	& 53 	& 22 	& 113 \\
		\textbf{Offline} 	& 1 	& \textbf{511} 	& 11 	& 0 	& 0 \\
		\textbf{Poor} 		& 67 	& 3 	& \textbf{779} 	& 31 	& 46 \\
		\textbf{Snow} 		& 36 	& 0 	& 35 	& \textbf{646} 	& 40 \\
		\textbf{Wet} 		& 117 	& 0 	& 8 	& 17 	& \textbf{781} \\
		\hline 
	\end{tabular}
	
\end{table}

\begin{description}
	\item [InceptionResNetV2:] This model was trained from the original ImageNet weights with all the layers set as trainable. There were a total of 54 million parameters to be trained. There were no modifications to the original model except for the final dense (output) layer. No dropout was applied.  Details are presented in Tables~\ref{tab:phase4-irn-classrep} and ~\ref{tab:phase4-irn-confmat}. 
\end{description}

\begin{table}[!htbp]
	\caption{\label{tab:phase4-irn-classrep}InceptionResNetV2 Classification report for validation set.}
	\centering{}
	\begin{tabular}{|c|cccc|c|}
		\hline 
		\textbf{IRNV2} & \textbf{Precision} & \textbf{Recall} & \textbf{F1-Score} & \textbf{Support} & \textbf{Accuracy}\\
		\hline 
		\textbf{Dry} 		& 0.90 & 0.92 & 0.91 & 1607 & \\
		\textbf{Offline} 	& 0.99 & 0.99 & 0.99 & 523 &\\
		\textbf{Poor} 		& 0.88 & 0.87 & 0.88 & 926 &\\
		\textbf{Snow} 		& 0.92 & 0.90 & 0.91 & 757 &\\
		\textbf{Wet} 		& 0.88 & 0.88 & 0.88 & 923 &\\
		\hline 
		\textbf{Training} 	&  &  &  &  & 90.8\%\\
		\textbf{Validation} &  &  &  &  & \textbf{90.7\%}\\
		\hline 
	\end{tabular}
	
\end{table}

\begin{table}[!htbp]
	\caption{\label{tab:phase4-irn-confmat} Confusion matrix for InceptionResNetV2 showing true labels
		vs predicted labels for validation.}
	\centering{}
	\begin{tabular}{|c|ccccc|}
		\hline 
		\textbf{T \textbackslash{} P} & \textbf{Dry} & \textbf{Offline} & \textbf{Poor} & \textbf{Snow} & \textbf{Wet}\\
		\hline 
		\textbf{Dry} 		& \textbf{1474} 	& 0 	& 67 	& 11 	& 55 \\
		\textbf{Offline} 	& 1 	& \textbf{519} 	& 3 	& 0 	& 0 \\
		\textbf{Poor} 		& 58 	& 4 	& \textbf{810} 	& 18 	& 36 \\
		\textbf{Snow} 		& 25 	& 0 	& 30 	& \textbf{684} 	& 18 \\
		\textbf{Wet} 		& 75 	& 0 	& 10 	& 30 	& \textbf{808} \\
		\hline 
	\end{tabular}
	
\end{table}

\begin{description}
	\item [EfficientNet-B4:] This model was trained for 4 epochs. Leaky ReLU was used for activation. The results are shown in tables \ref{tab:phase4-ef4-classrep} and \ref{tab:phase4-ef4-confmat}. 
\end{description}

\begin{table}[!htbp]
	\caption{\label{tab:phase4-ef4-classrep}EfficientNet-B4 classification report for validation set.}
	\centering{}
	\begin{tabular}{|c|cccc|c|}
		\hline 
		\textbf{EN-B4} & \textbf{Precision} & \textbf{Recall} & \textbf{F1-Score} & \textbf{Support} & \textbf{Accuracy}\\
		\hline 
		\textbf{Dry} 		& 0.90 & 0.92 & 0.91 & 1607 & \\
		\textbf{Offline} 	& 0.98 & 0.99 & 0.99 & 523 &\\
		\textbf{Poor} 		& 0.91 & 0.84 & 0.87 & 926 &\\
		\textbf{Snow} 		& 0.94 & 0.92 & 0.93 & 757 &\\
		\textbf{Wet} 		& 0.86 & 0.92 & 0.89 & 923 &\\
		\hline 
		\textbf{Training} 	&  &  &  &  & 90.3\%\\
		\textbf{Validation} &  &  &  &  & \textbf{90.9\%}\\
		\hline 
	\end{tabular}
	
\end{table}

\begin{table}[!htbp]
	\caption{\label{tab:phase4-ef4-confmat} Confusion matrix for EfficientNet-B4 showing true labels
		vs predicted labels for validation.}
	\centering{}
	\begin{tabular}{|c|ccccc|}
		\hline 
		\textbf{T \textbackslash{} P} & \textbf{Dry} & \textbf{Offline} & \textbf{Poor} & \textbf{Snow} & \textbf{Wet}\\
		\hline 
		\textbf{Dry} 		& \textbf{1471} 	& 0 	& 56 	& 7 	& 73 \\
		\textbf{Offline} 	& 0 	& \textbf{520} 	& 3 	& 0 	& 0 \\
		\textbf{Poor} 		& 75 	& 8 	& \textbf{776} 	& 23 	& 44 \\
		\textbf{Snow} 		& 29 	& 1 	& 17 	& \textbf{693} 	& 17 \\
		\textbf{Wet} 		& 54 	& 0 	& 5 	& 18 	& \textbf{846} \\
		\hline 
	\end{tabular}
	
\end{table}

\subsubsection{Analysis}

With 87.3\% validation accuracy over 5 classes, VGG-16 maintained a slightly worse validation accuracy on the 47K set than the previous 20K cherry-picked data set. This is noteworthy since this set was much more representative of the unlabelled data set as half of it consisted of unrestricted (non-cherry-picked) images and the other half is strictly cherry-picked images. Also note that the gap between training and validation accuracies is less than 2\% showing we are not over-fitting the training data. InceptionResNetV2 achieved an even higher accuracy. Within 4 epochs, it reached 90.7\%, beating even the cherry-picked experiments of VGG-16 in Section \ref{sec:five-class}. This model also did not suffer from over-fitting as the training accuracy was only 0.1\% better than the validation accuracy. EfficientNet-B4 performed slightly better than InceptionResNetV2. After epoch 4, it achieved 90.9\% validation accuracy overall, yielding the highest accuracy. It achieved the highest F-1 score for wet and snow, and tied with InceptionResNetV2 for dry and poor classes.

To observe their performance on the unlabelled set, again we labelled the random extract with 1000 images from the 1.5 million data set, described in Section ~\ref{sec:five-class-diverse}. Table~\ref{tab:random1000-47K} shows the combined results. For the frameworks trained on the 47K data set, we can see that the number of false positives decreased across all classes. There were also significant improvements for snow and wet classes, compared to the VGG-16 classification on the cherry-picked 20K set. Once again, EfficientNet-B4 proved to the most accurate framework. It achieved the lowest false positives on dry, snow and wet classes. It also achieved the highest true positives on the wet class.

\begin{table}[!htbp]
	\caption{\label{tab:random1000-47K}Summary of classification evaluation for 1000 random images from the combined (1.5M) data set.}
	
	\centering{}%
	\begin{tabular}{|c|cccccc|c|}
		\hline 
		& \textbf{Verdict} & \textbf{Dry} & \textbf{Offline} & \textbf{Poor} & \textbf{Snow} & \textbf{Wet} & \textbf{Total (1000)}\\
		\hline 
		\textbf{VGG-16} & Acceptable: & 616 & 65 & 136 & 9 & 62 & 888\\
		\textbf{20K} & \textbf{Refused:} & \textbf{31} & \textbf{0} & \textbf{7} & \textbf{21} & \textbf{53} & \textbf{112}\\
		\hline 
		\textbf{VGG-16} & Acceptable: & 599 & 79 & 205 & 8 & 70 & 961\\
		\textbf{47K} & \textbf{Refused:} & \textbf{17} & \textbf{0} & \textbf{0} & \textbf{1} & \textbf{21} & \textbf{39}\\
		\hline 
		\textbf{IRN-V2} & Acceptable: & 587 & 76 & 247 & 10 & 54 & 974\\
		\textbf{47K} & \textbf{Refused:} & \textbf{20} & \textbf{0} & \textbf{0} & \textbf{3} & \textbf{3} & \textbf{26}\\
		\hline 
		\textbf{EN-B4} & Acceptable: & 608 & 78 & 217 & 9 & 66 & 978\\
		\textbf{47K} & \textbf{Refused:} & \textbf{18} & \textbf{1} & \textbf{0} & \textbf{1} & \textbf{2} & \textbf{22}\\
		\hline 
	\end{tabular}
\end{table}

Training with more data, diversifying the poor class and using a higher-end classifier with more layers trained reflected on the results very positively. The number of evident false positives significantly decreased overall. In the next section, we'll perform a comparative analysis of the run-time performances for multiple frameworks.

\section{Classification Results and Real-time Map Building}
\label{sec:finalresults}
In this section, we give comparative results from trained 6 deep-learning models: VGG-16, ResNet50, Xception, InceptionResNetV2, EfficientNet-B0 and EfficientNet-B4. They were trained from scratch, on the 90-10 split 47K data set to see how they would compare as far as accuracy and execution time (training + validation) was concerned. Table \ref{tab:Common-configurations} shows the common configurations and the hyper-parameters that were shared across the frameworks. Table \ref{tab:Algorithm-specific-configuration} shows the algorithm-specific configurations that were shaped using heuristics or default/suggested values. For the first 4 algorithms, we used the official Keras implementations and applied transfer learning as usual. However, we used the following implementation of EfficientNet\footnote{https://github.com/qubvel/efficientnet}.

\begin{table}[!htbp]
	\caption{\label{tab:Common-configurations}Common configurations}
	
	\centering{}%
	\begin{tabular}{|rl|}
		\hline 
		\textbf{Hyper-parameter} & \textbf{: Value}\\
		\hline 
		\hline 
		Optimizer & : Rectified Adam\\
		\hline 
		Learning Rate & : 0.0001\\
		\hline 
		Loss Func. & : Categorical Crossentropy\\
		\hline 
		Activation Func. & : Softmax\\
		\hline 
		Global Pooling & : Average\\
		\hline 
		Batch Size & : 16\\
		\hline 
		Max Epochs & : 12\\
		\hline 
		0-1 Rescaling & : Yes\\
		\hline 
		Training Set & : 42606\\
		\hline 
		Validation Set & : 4736\\
		\hline 
		Target Classes & : 5\\
		\hline 
		Augmented Training & : Yes\\
		\hline 
	\end{tabular}
\end{table}

\begin{figure}[!h]
	\centering{}
		\includegraphics[width=0.35\textwidth]{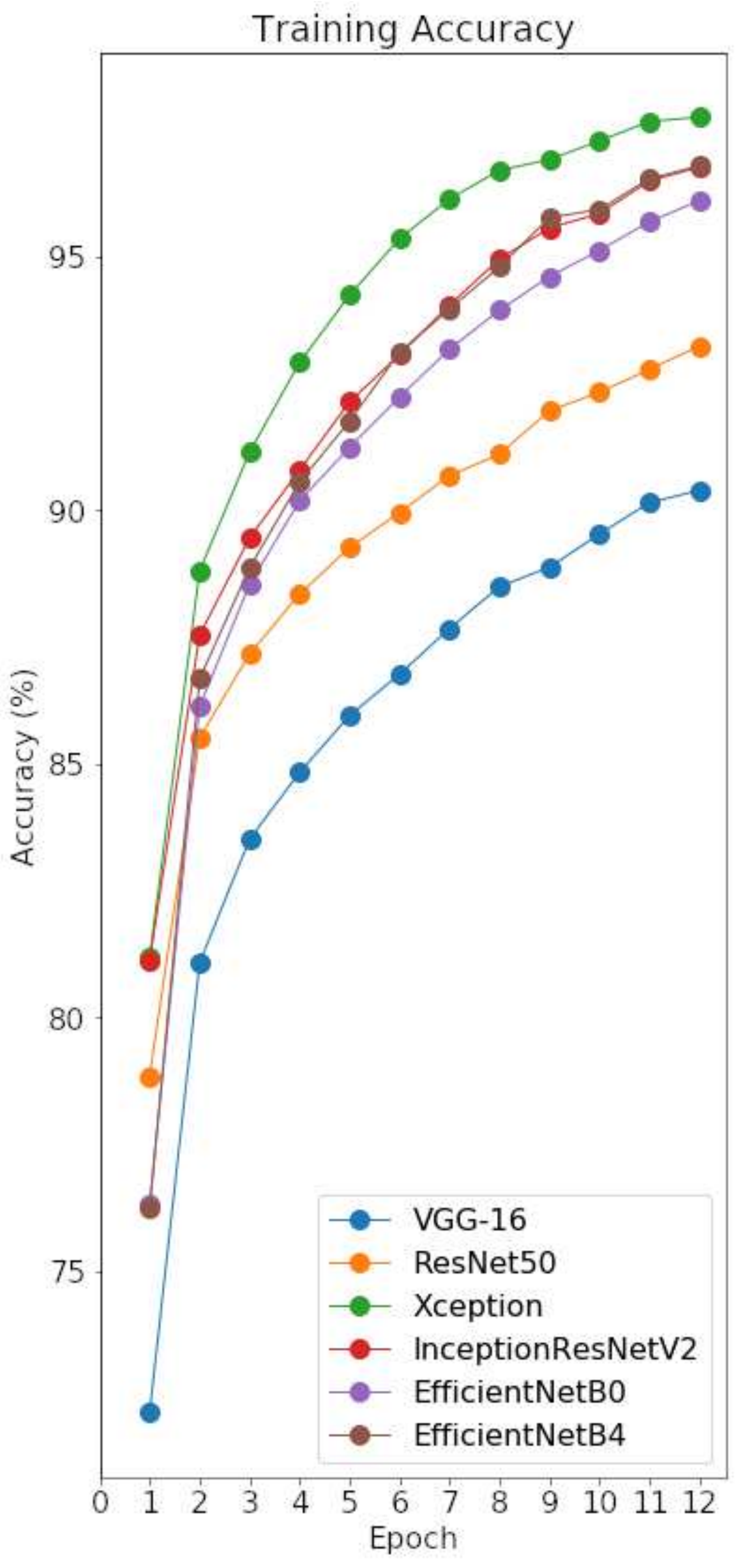}\includegraphics[width=0.35\textwidth]{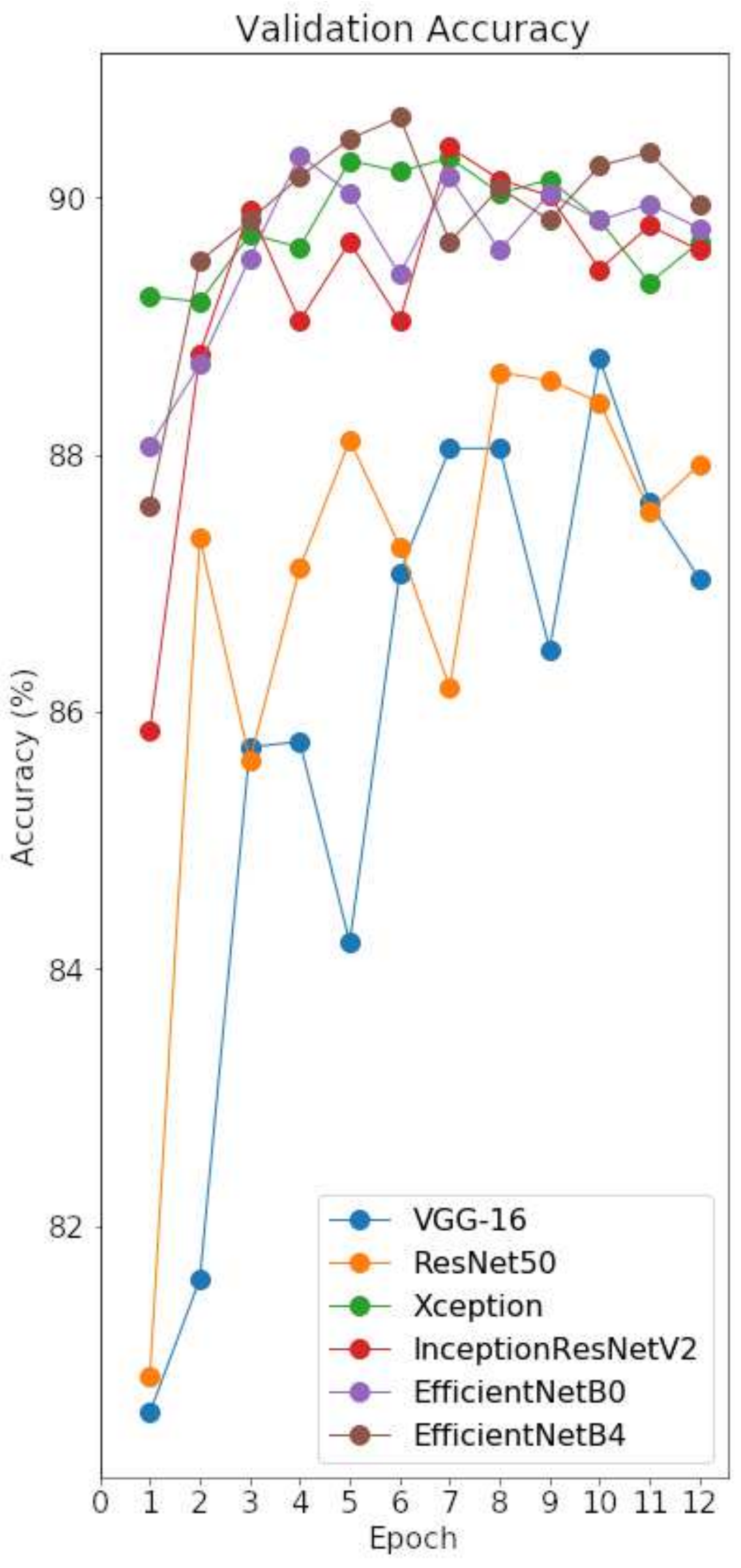}
	\caption{\label{fig:Training-=000026-Validation}Training \& Validation accuracies
		over 12 epochs for 6 algorithms}
\end{figure}

\begin{figure}[!h]
	\centering{}
		\includegraphics[width=.5\textwidth]{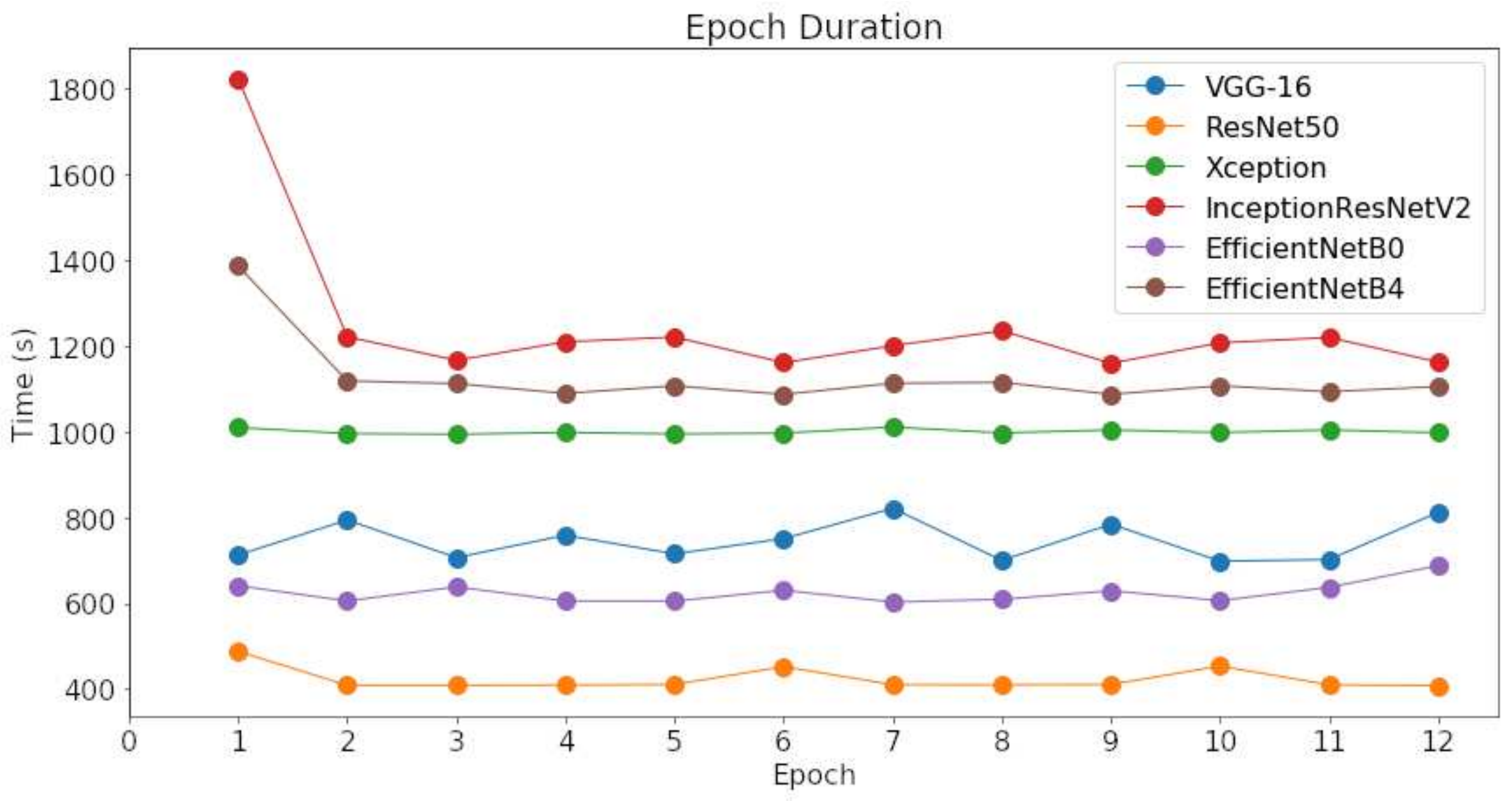}
		\includegraphics[width=.5\textwidth]{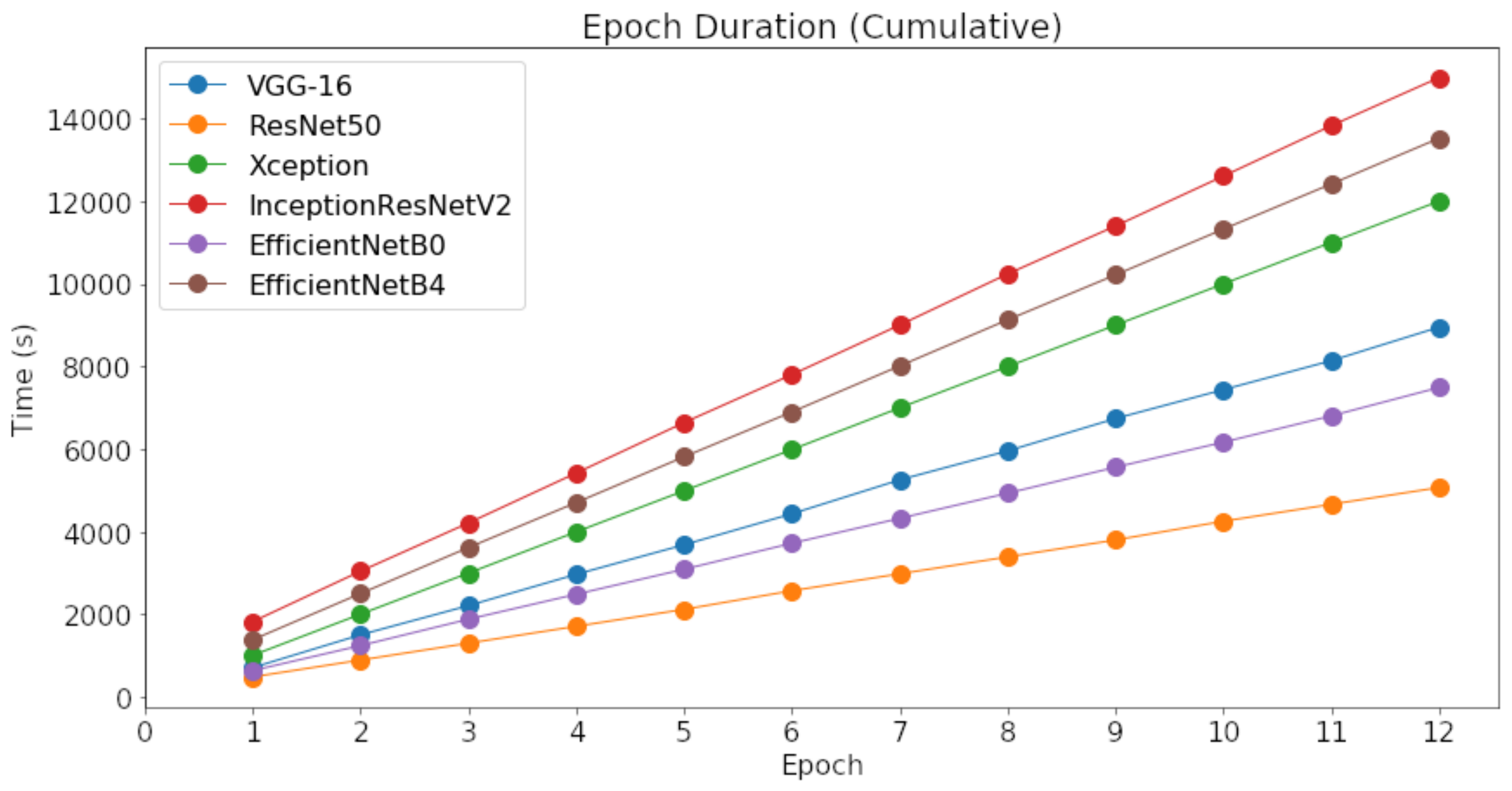}
		
	\caption{\label{fig:Cumulative-epoch-durations}Execution times (training +
		validation + model saving) over 12 epochs for 6 algorithms. Top: Per
		epoch. Bottom: Cumulative}
\end{figure}

Accuracies and the execution times of these experiments can be seen in Figures~\ref{fig:Training-=000026-Validation} and ~\ref{fig:Cumulative-epoch-durations} respectively. As far as the accuracy is concerned, we can see that VGG-16 and ResNet lag behind the XCeption, InceptionResNetV2 frameworks and the most recent EfficientNet frameworks compete well. In this particular set of runs, the highest observed validation accuracy was 90.6\% after epoch 6 of EfficientNet-B4 (1200ms execution time) while the others take turns to achieve the highest validation accuracy in other epochs. For the execution times, EfficientNet-B0 is the winner (600ms execution time). It is only bested by ResNet50 (400ms execition time) which nonetheless has poorer validation accuracy (85.67\%). 

\begin{table}[!h]
	\caption{\label{tab:Algorithm-specific-configuration}Algorithm-specific configurations.
		H: Hidden. D: Dropout.}
	
	\centering{}\makebox[\textwidth][c]{%
	\begin{tabular}{|r|cccccc|}
		\hline 
		& \textbf{VGG-16} & \textbf{RN-50} & \textbf{XCep.} & \textbf{IRN-V2} & \textbf{EN-B0} & \textbf{EN-B4}\\
		\hline 
		\hline 
		\textbf{Input Dims} & (224, 224) & (224, 224) & (299,299) & (299, 299) & (256, 256) & (256, 256)\\
		\hline 
		\textbf{Global Pooling} & None & Average & Average & Average & Average & Average\\
		\hline 
		\textbf{Top Layers} & H-D-H-D & D & D & D & D & D\\
		\hline 
		\textbf{FC Neurons} & 4096 & 0 & 0 & 0 & 0 & 0\\
		\hline 
		\textbf{Dropout Rate} & 0.4 & 0.4 & 0.2 & 0.4 & 0.2 & 0.4\\
		\hline 
		\textbf{Total Layers} & 25 & 178 & 135 & 783 & 233 & 470\\
		\hline 
		\textbf{Total Prms.} & 134M & 23M & 20M & 54M & 4M & 18M\\
		\hline 
		\textbf{Trainable Prms} & 134M & 23M & 20M & 54M & 4M & 18M\\
		\hline 
		\textbf{Nontrain. Prms} & 0 & 53K & 54K & 60K & 42K & 125K\\
		\hline 
	\end{tabular}}
\end{table}

EfficientNet-B0 takes about half the time of its competitors, suggesting that it may be a good trade-off. Its highest validation accuracy was 90.3\%. VGG-16 with transfer learning proved to be very useful for data acquisition and pseudo-labelling with limited hardware resources, throughout this project. However, it seems more recent frameworks like XCeption, InceptionResNetV2 and EfficientNet performed better and they are all shown to be good candidates for our problem, provided that the hardware to handle them is available.

\begin{figure}[!h]
	\includegraphics[width=10cm]{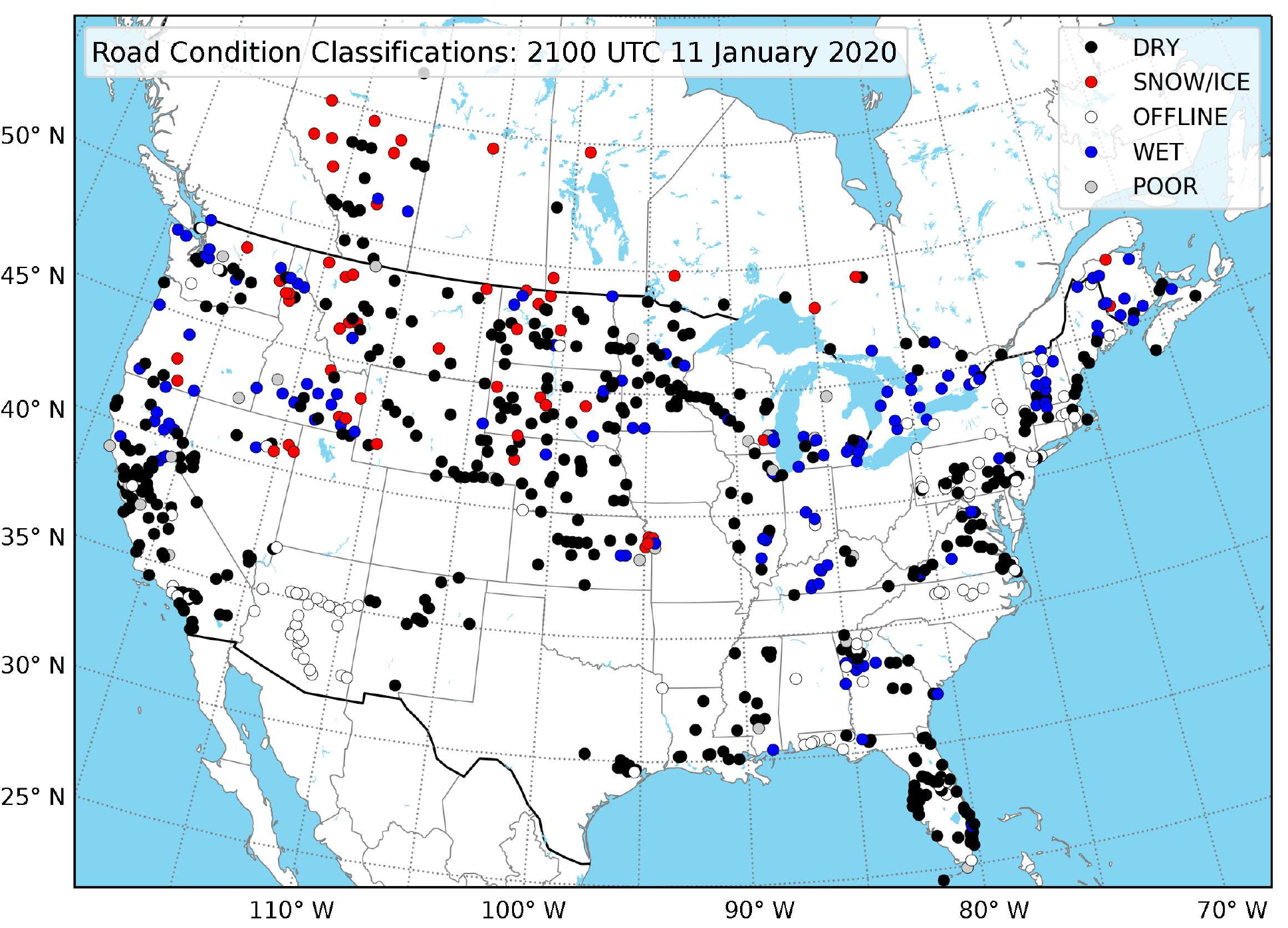}
	\centering
	\caption{An example map of 782 classified camera images over Canada and the United States at 2100 UTC 11 January 2020. Each marker represents one classified image with the legend indicating the colour corresponding to each respective class}
	\label{fig:rtmap}
\end{figure}

Once all images in the pipeline have been classified, the data is stored both in csv files and can also be uploaded to a PostgreSQL database, if desired. These output data contain the image name, latitude and longitude, and class. A map plotting program then takes the data and produces a map for the desired domain. An example output map for all of North America is shown in Figure~\ref{fig:rtmap} with 782 classified images using the EfficientNetB4 framework.

\section{Conclusion}
\label{sec:con}

In this paper, we have presented a detailed account of the process of generating training examples of images from live camera feeds depicting a wide range of road and weather conditions in North America.  Weather understanding is an important part of many real-world applications that involves transportation and road safety. Our process involved leveraging deep convolutional neural networks in conjunction with manual labelling to produce reasonable quality training examples. The proposed application pipeline includes a map building component which is the one of the main outcomes of this research.    We demonstrated that recent deep convolutional neural networks were able to produce good results with a maximum accuracy of 90.9\% without any pre-processing of the images.  The choice of these frameworks and our analysis take into account unique requirements of real-time map building functions. So, in addition to the classification accuracy,  model complexity and memory usage must be taken into account during the different stages of dataset labelling.
 
Our experiments show that with an increasing number of training examples available with more diverse content, the results tend to improve. With sufficiently large training set, the performance of these frameworks on other benchmarks also seemed to reflect well into this particular problem, with newer ones with higher benchmark scores in other problems also performing better~\cite{Bianco18}. 

Future research directions include experimenting with other resource demanding frameworks with good potential such as NASNet-Large~\cite{nasnet} and EfficientNet-B7~\cite{enet} on more advanced hardware. In addition, we will consider employing ensemble learning techniques~\cite{ensemble_methods} in order to exploit the best aspects of multiple frameworks. We plan to seek the possibilities to incorporate road detection and segmentation~\cite{road_segmentation1, road_segmentation2} as a preprocessing step to crop the images and focus on the road segments. This might be especially helpful on images with low visibility of the roads such as cameras placed far away from the roads.

\bibliographystyle{elsart-num}
\bibliography{./DNN}

\begin{thebibliography}{10}
\expandafter\ifx\csname url\endcsname\relax
  \def\url#1{\texttt{#1}}\fi
\expandafter\ifx\csname urlprefix\endcsname\relax\def\urlprefix{URL }\fi

\bibitem{crevier2001}
L.-P. Crevier, Y.~Delage, {METRo: A New Model for Road-Condition Forecasting in
  Canada}, Journal of Applied Meteorology 40~(11) (2001) 2026--2037.
\newline\urlprefix\url{https://doi.org/10.1175/1520-0450(2001)040<2026:MANMFR>2.0.CO;2}

\bibitem{Sass1997}
B.~H. Sass, {A Numerical Forecasting System for the Prediction of Slippery
  Roads}, Journal of Applied Meteorology 36~(6) (1997) 801--817.
\newline\urlprefix\url{https://doi.org/10.1175/1520-0450(1997)036<0801:ANFSFT>2.0.CO;2}

\bibitem{Drobot2014}
S.~Drobot, A.~R.~S. Anderson, C.~Burghardt, P.~Pisano, U.s. public preferences
  for weather and road condition information, Bulletin of the American
  Meteorological Society 95~(6) (2014) 849--859.
\newline\urlprefix\url{https://doi.org/10.1175/BAMS-D-12-00112.1}

\bibitem{Carillo2019}
J.~Carrillo, M.~Crowley, G.~Pan, L.~Fu, {Comparison of Deep Learning models for
  Determining Road Surface Condition from Roadside Camera Images and Weather
  Data}, in: Transportation Association of Canada and Intelligent
  Transportation Systems Canada Joint Conference, 2019, pp. 1--16.

\bibitem{KurihataRain2005}
H.~Kurihata, T.~Takahashi, I.~Ide, Y.~Mekada, H.Murase, Y.~Tamatsu,
  T.~Miyahara, Rainy weather recognition from in-vehicle camera images for
  driver assistance, in: IEEE Proceedings, Intelligent Vehicles Symposium,
  2005, pp. 205--210.

\bibitem{Hautiere2006}
N.~Hautiere, J.-P. Tarel, J.~Lavenant, D.~Aubert, Automatic fog detection and
  estimation of visibility distance through use of an onboard camera, Mach.
  Vision Appl. 17~(1) (2006) 8--20.

\bibitem{RoserRain2008}
M.~Roser, F.~Moosmann, Classification of weather situations on single color
  images, in: IEEE Proceedings, Intelligent Vehicles Symposium, 2008, pp.
  798--803.

\bibitem{Yan2009}
X.~Yan, Y.~Luo, X.~Zheng, Weather recognition based on images captured by
  vision system in vehicle, in: Proceedings of the 6th International Symposium
  on Neural Networks: Advances in Neural Networks - Part III, 2009, pp.
  390--398.

\bibitem{Bronte2009FogDS}
S.~Bronte, L.~M. Bergasa, P.~F. Alcantarilla, Fog detection system based on
  computer vision techniques, in: 2009 12th International IEEE Conference on
  Intelligent Transportation Systems, 2009, pp. 1--6.

\bibitem{Omer2010AnAI}
R.~Omer, L.~Fu, An automatic image recognition system for winter road surface
  condition classification, 13th International IEEE Conference on Intelligent
  Transportation Systems (2010) 1375--1379.

\bibitem{Gallen2011}
N.~H. R.~Gallen, A.~Cord, D.~Aubert, Towards night fog detection through use of
  in-vehicle multipurpose cameras, in: Proceedings of the 2011 IEEE Intelligent
  Vehicles Symposium (IV), 2011, pp. 399--404.

\bibitem{Pavli2012}
M.~Pavli, H.~Belzner, G.~Rigoll, S.~Ili, Image based fog detection in vehicles,
  in: Proceedings of the 2012 IEEE Intelligent Vehicles Symposium (IV), 2012,
  pp. 1132--1137.

\bibitem{Zhang2015MulticlassWC}
Z.~Zhang, H.~Ma, Multi-class weather classification on single images, in: 2015
  IEEE International Conference on Image Processing (ICIP), 2015, pp.
  4396--4400.

\bibitem{Almazan2016}
E.~J. Almazan, Y.~Qian, J.~H. Elder, Road segmentation for classification of
  road weather conditions, in: G.~Hua, H.~J{\'e}gou (Eds.), Computer Vision --
  ECCV 2016 Workshops, Springer International Publishing, Cham, 2016, pp.
  96--108.

\bibitem{Amtor2015}
M.~Amthor, B.~Hartmann, J.~Denzler, Road condition estimation based on
  spatio-temporal reflection models, in: J.~Gall, P.~Gehler, B.~Leibe (Eds.),
  Pattern Recognition, Springer International Publishing, Cham, 2015, pp.
  3--15.

\bibitem{Narasimhan2003}
S.~G. Narasimhan, S.~K. Nayar, Shedding light on the weather, in: IEEE
  Conference on Computer Vision and Pattern Recognition (CVPR), 2003, pp. I--I.

\bibitem{LICVPR2009}
L.~Shen, Photometric stereo and weather estimation using internet images, in:
  IEEE Conference on Computer Vision and Pattern Recognition (CVPR), 2009, pp.
  1850--1857.

\bibitem{Laffont2014}
P.-Y. Laffont, Z.~Ren, X.~Tao, C.~Qian, J.~Hays, Transient attributes for
  high-level understanding and editing of outdoor scenes, ACM Transactions on
  Graphics 33~(4) (2014) 149:1--149:11.

\bibitem{Hong2014}
H.~Song, Y.~Chen, Y.~Gao, Weather condition recognition based on feature
  extraction and k-nn, in: F.~Sun, D.~Hu, H.~Liu (Eds.), Foundations and
  Practical Applications of Cognitive Systems and Information Processing,
  Springer Berlin Heidelberg, Berlin, Heidelberg, 2014, pp. 199--210.

\bibitem{Li2014}
Q.~Li, Y.~Kong, S.~M. Xia, A method of weather recognition based on outdoor
  images, in: Computer Vision Theory and Applications (VISAPP), 2014
  International Conference, 2014, pp. 510--516.

\bibitem{schmidhuber2015deep}
J.~Schmidhuber, Deep learning in neural networks: An overview, Neural Networks
  61 (2015) 85--117.

\bibitem{lecun2015deep}
Y.~LeCun, Y.~Bengio, G.~Hinton, Deep learning, Nature 521~(7553) (2015) 436.

\bibitem{Krizhevsky2012}
A.~Krizhevsky, I.~Sutskever, G.~E. Hinton, Imagenet classification with deep
  convolutional neural networks, in: Advances in Neural Information Processing
  Systems 25, 2012, pp. 1097--1105.

\bibitem{Russakovsky2015}
O.~Russakovsky, J.~Deng, H.~Su, J.~Krause, S.~Satheesh, S.~Ma, Z.~Huang,
  A.~Karpathy, A.~Khosla, M.~Bernstein, A.~C. Berg, L.~Fei-Fei, {ImageNet Large
  Scale Visual Recognition Challenge}, International Journal of Computer Vision
  (IJCV) 115~(3) (2015) 211--252.

\bibitem{LiuPAMI2017}
C.~{Lu}, D.~{Lin}, J.~{Jia}, C.~{Tang}, Two-class weather classification, IEEE
  Transactions on Pattern Analysis and Machine Intelligence 39~(12) (2017)
  2510--2524.

\bibitem{RSCMLin2017}
D.~{Lin}, C.~{Lu}, H.~{Huang}, J.~{Jia}, Rscm: Region selection and concurrency
  model for multi-class weather recognition, IEEE Transactions on Image
  Processing 26~(9) (2017) 4154--4167.

\bibitem{zhao2019cnnrnn}
B.~Zhao, X.~Li, X.~Lu, Z.~Wang, A cnn-rnn architecture for multi-label weather
  recognition (2019).

\bibitem{Elhoseiny2015WeatherCW}
M.~Elhoseiny, S.~Huang, A.~M. Elgammal, Weather classification with deep
  convolutional neural networks, 2015 IEEE International Conference on Image
  Processing (ICIP) (2015) 3349--3353.

\bibitem{Zhu2016ExtremeWR}
Z.~Zhu, L.~Zhuo, P.~Qu, K.~Zhou, J.~Zhang, Extreme weather recognition using
  convolutional neural networks, 2016 IEEE International Symposium on
  Multimedia (ISM) (2016) 621--625.

\bibitem{Guerra2018}
J.~C. {Villarreal Guerra}, Z.~{Khanam}, S.~{Ehsan}, R.~{Stolkin},
  K.~{McDonald-Maier}, Weather classification: A new multi-class dataset, data
  augmentation approach and comprehensive evaluations of convolutional neural
  networks, in: 2018 NASA/ESA Conference on Adaptive Hardware and Systems
  (AHS), 2018, pp. 305--310.

\bibitem{LiGAN2018}
Z.~{Li}, Y.~{Jin}, Y.~{Li}, Z.~{Lin}, S.~{Wang}, Imbalanced adversarial
  learning for weather image generation and classification, in: 2018 14th IEEE
  International Conference on Signal Processing (ICSP), 2018, pp. 1093--1097.

\bibitem{Pan2018RoadSurface}
G.~Pan, L.~Fu, R.~Yu, M.~Muresan, Winter road surface condition recognition
  using a pre-trained deep convolutional neural network, arXiv preprint
  abs/1812.06858.
\newline\urlprefix\url{http://arxiv.org/abs/1812.06858}

\bibitem{Nolte2018RoadSurface}
M.~Nolte, N.~Kister, M.~Maurer, Assessment of deep convolutional neural
  networks for road surface classification, arXiv preprint abs/1812.08872.
\newline\urlprefix\url{https://arxiv.org/abs/1804.08872}

\bibitem{simonyan2014very}
K.~Simonyan, A.~Zisserman, {Very Deep Convolutional Networks for Large-Scale
  Image Recognition}, arXiv preprint arXiv:1409.1556.

\bibitem{He2016}
K.~He, X.~Zhang, S.~Ren, J.~Sun, {Deep Residual Learning for Image
  Recognition}, in: 2016 IEEE Conference on Computer Vision and Pattern
  Recognition (CVPR), 2016, pp. 770--778.

\bibitem{szegedy2017inception}
C.~Szegedy, S.~Ioffe, V.~Vanhoucke, A.~Alemi, {Inception-v4, Inception-ResNet
  and the Impact of Residual Connections on Learning}, in: AAAI, Vol.~4, 2016,
  p.~12.

\bibitem{xception}
F.~Chollet, Xception: Deep learning with depthwise separable convolutions, in:
  2017 IEEE Conference on Computer Vision and Pattern Recognition (CVPR), 2017,
  pp. 1800--1807.

\bibitem{enet}
M.~Tan, Q.~V. Le, Efficientnet: Rethinking model scaling for convolutional
  neural networks, CoRR abs/1905.11946.
\newline\urlprefix\url{http://arxiv.org/abs/1905.11946}

\bibitem{Bianco18}
S.~Bianco, R.~Cad{\`{e}}ne, L.~Celona, P.~Napoletano, Benchmark analysis of
  representative deep neural network architectures, CoRR abs/1810.00736.
\newline\urlprefix\url{http://arxiv.org/abs/1810.00736}

\bibitem{nasnet}
B.~Zoph, V.~Vasudevan, J.~Shlens, Q.~V. Le, Learning transferable architectures
  for scalable image recognition, CoRR abs/1707.07012.
\newline\urlprefix\url{http://arxiv.org/abs/1707.07012}

\bibitem{ensemble_methods}
R.~Maclin, D.~W. Opitz, Popular ensemble methods: An empirical study, CoRR
  abs/1106.0257.
\newline\urlprefix\url{http://arxiv.org/abs/1106.0257}

\bibitem{road_segmentation1}
Y.~Lyu, X.~Huang, Road segmentation using {CNN} with {GRU}, CoRR
  abs/1804.05164.
\newline\urlprefix\url{http://arxiv.org/abs/1804.05164}

\bibitem{road_segmentation2}
P.~Chen, H.~Hang, S.~Chan, J.~Lin, Dsnet: An efficient {CNN} for road scene
  segmentation, CoRR abs/1904.05022.
\newline\urlprefix\url{http://arxiv.org/abs/1904.05022}

\end{thebibliography}

\end{document}